\documentclass{article}

\usepackage{microtype}
\usepackage{graphicx}
\usepackage{subfigure}
\usepackage{booktabs} %
\usepackage{enumitem}
\usepackage[dvipsnames]{xcolor}
\usepackage{listings}
\usepackage{lstautogobble}  %

\usepackage{hyperref}

\usepackage[accepted]{icml2026}

\usepackage{amsmath}
\usepackage{amssymb}
\usepackage{mathtools}
\mathtoolsset{showonlyrefs}
\usepackage{amsthm}
\usepackage{bm}
\usepackage{xspace}

\usepackage{mdframed}
\definecolor{theoremcolor}{rgb}{255, 255, 255}
\mdfsetup{
    backgroundcolor=theoremcolor,
    linewidth=1pt,
    leftline=true,
    topline=false,
    rightline=false,
    bottomline=false 
}

\newmdtheoremenv{definition}{Definition}
\newmdtheoremenv{proposition}{Proposition}
\newmdtheoremenv{corollary}{Corollary}
\newmdtheoremenv{theorem}{Theorem}
\newmdtheoremenv{lemma}{Lemma}
\newmdtheoremenv{example}{Example}

\def\h{{\bm{h}}}
\def\s{{\bm{s}}}
\def\x{{\bm{x}}}
\def\y{{\bm{y}}}
\def\z{{\bm{z}}}
\def\tauv{{\bm{\tau}}}

\def\RR{{\mathbb{R}}}
\def\NN{{\mathbb{N}}}
\def\EE{{\mathbb{E}}}
\def\PP{{\mathbb{P}}}

\def\cA{{\mathcal{A}}}

\def\cF{{\mathcal{F}}}
\def\cL{{\mathcal{L}}}
\def\cM{{\mathcal{M}}}
\def\cP{{\mathcal{P}}}
\def\cS{{\mathcal{S}}}
\def\cT{{\mathcal{T}}}

\def\cV{{\mathcal{V}}}
\def\cX{{\mathcal{X}}}
\def\cY{{\mathcal{Y}}}

\newcommand{\Qsub}[1]{R_{#1}}
\newcommand{\Rsub}[1]{R_{#1}}

\def\arm{{\textsc{arm}}}
\def\ebm{{\textsc{ebm}}}
\def\lse{{\mathrm{LSE}}}
\def\softargmax{{\operatorname{softargmax}}}
\def\KL{{\mathrm{KL}}}
\def\ones{{\mathbf{1}}}

\def\len{{\mathrm{len}}}
\def\bos{\texttt{BOS}\xspace}
\def\eos{\texttt{EOS}\xspace}

\def\voc{\cV}
\def\augvoc{\cA}

\def\iid{{i.i.d.\ }}
\def\aka{{a.k.a.\ }}
\def\wrt{{w.r.t.\ }}

\def\pref{p_{\mathrm{ref}}}
\def\piref{\pi_{\mathrm{ref}}}

\DeclareMathOperator*{\argmax}{argmax}
\DeclareMathOperator*{\argmin}{argmin}

\def\arsim{\overset{\text{AR}}{\sim}}

\definecolor{bluekeywords}{rgb}{0.13, 0.13, 1}
\definecolor{greencomments}{rgb}{0, 0.5, 0}
\definecolor{redstrings}{rgb}{0.9, 0, 0}
\definecolor{graynumbers}{rgb}{0.5, 0.5, 0.5}

\usepackage[textsize=tiny]{todonotes}

\def\mytitle{Autoregressive Language Models are Secretly Energy-Based Models:\\
Insights into the Lookahead Capabilities of Next-Token Prediction}

\icmltitlerunning{Autoregressive Language Models are Secretly Energy-Based Models}

\begin{document}

\twocolumn[
\icmltitle{\mytitle}

\icmlsetsymbol{equal}{*}

\begin{icmlauthorlist}

\icmlauthor{Mathieu  Blondel}{gdm}
\icmlauthor{Michaël E. Sander}{gdm}
\icmlauthor{Germain Vivier-Ardisson}{gdm}
\icmlauthor{Tianlin Liu}{gdm}
\icmlauthor{Vincent Roulet}{gdm}

\end{icmlauthorlist}

\icmlaffiliation{gdm}{Google DeepMind}

\icmlcorrespondingauthor{}{mblondel@google.com}

\icmlkeywords{autoregressive models, energy-based model, maxent RL, dynamic programming}

\vskip 0.3in
]

\printAffiliationsAndNotice{}  %

\begin{abstract}
Autoregressive models (ARMs) currently constitute the dominant paradigm for
large language models (LLMs). Energy-based models (EBMs) represent another class
of models, which have historically been less prevalent in LLM development, yet
naturally characterize the optimal policy in post-training alignment. In this
paper, we present a unified view of these two model classes. Taking the chain
rule of probability as a starting point, we establish an explicit bijection
between ARMs and EBMs in function space, which we show to correspond to a
special case of the soft Bellman equation in maximum entropy reinforcement
learning.
Building upon this bijection, we derive the equivalence between
supervised learning of ARMs and EBMs. Furthermore, we analyze the
distillation of EBMs into ARMs by providing theoretical error bounds. 
Our results provide insights into the ability of ARMs to plan ahead, despite
being based on the next-token prediction paradigm.
\end{abstract}

\section{Introduction}

Autoregressive models (ARMs), which are based on the next-token prediction
paradigm, currently constitute the dominant approach for
large language models (LLMs). 
Their
primary advantages lie in their highly parallelizable training and the ability
to perform exact ancestral sampling.
Energy-based models (EBMs) represent another class of models,
which inherently have the ability to look ahead,
since they define sequence-level
distributions. However, they have historically been less prevalent, since they
require computing an intractable partition function over the set of all possible
sequences. Consequently, they are significantly more challenging to train and
sample from, often necessitating Markov-chain Monte-Carlo (MCMC) methods.

Following pre-training, LLMs typically undergo a post-training alignment phase.
This process is frequently framed as maximum entropy (MaxEnt) reinforcement
learning (RL), \aka entropy-regularized RL, where we seek a probability
distribution (or policy) that maximizes a trade-off between the expected reward
and a Kullback-Leibler (KL) regularization term. Crucially, if we maximize over
the set of all possible probability distributions, the analytical solution to
this objective is exactly an EBM.
Therefore, finding an ARM that maximizes
this objective is equivalent 
to distilling an EBM into an ARM.

This perspective raises a fundamental question: To what extent can ARMs
approximate EBMs?  From a probabilistic graphical model standpoint, ARMs are
Bayesian networks (directed graphical models), whereas EBMs are Markov random
fields (undirected graphical models).  However, ARMs rely on the chain rule of
probability, which can, in principle, decompose any probability distribution.
It naturally follows that ARMs should be convertible into EBMs, and vice-versa.
Building upon these intuitions, we derive several results that provide insights
into the approximation power of ARMs \wrt EBMs, and consequently
their ability to plan ahead, despite being based on the next-token prediction
paradigm.

\paragraph{Contributions.}

\begin{itemize}[topsep=0pt,itemsep=3pt,parsep=3pt,leftmargin=5pt]

\item We provide a unified view of ARMs and EBMs, exhibiting clear structural
    analogies between these two model classes.

\item In function space (\aka tabular setting in RL), we
    establish an exact bijection between ARMs and EBMs (Proposition \ref{prop:bijection}). 
    While our starting point
    is the chain rule of probability, our bijection can be viewed as an instance
    of the soft Bellman equation in MaxEnt RL. Our result demonstrates that in
    our setting the
    soft Bellman equation naturally emerges as a direct consequence of the chain
    rule of probability.

\item Building upon this bijection, we derive the equivalence
    between supervised learning of ARMs and EBMs (Proposition \ref{prop:same_minima}). Furthermore, we
    derive error bounds for distilling EBMs into ARMs (Proposition \ref{prop:kl_bound}), 
    a process identical to MaxEnt RL. 
    Our results, validated through numerical experiments,
    provide some theoretical justification for the next-token prediction
    paradigm and for teacher forcing.

\end{itemize}

\paragraph{Notation.}

Let $V$ be the vocabulary size, and let $\voc \coloneqq \{1, \ldots, V\}$ be the
vocabulary set.  We denote the set of sequences of size $t$ in the vocabulary by
$\voc^t$.  We consider two special tokens that are excluded from $\voc$: a \bos
(beginning of sequence) token that encodes the beginning of a prompt and a \eos
(end of sequence) token that encodes the end of a response.
We then denote the set of prompts of maximal size $U$ (\bos included) by $\cX =
\{\bos\} \times \cup_{t=1}^{U-1} \voc^t$,
and the set of responses of maximal size $T$ (\eos included) by 
$\cY = \cup_{t=1}^{T-1} \voc^t \times \{\eos\}$. 
A response $\y = (y_1, \dots, y_\tau) \in \cY$ always ends with \eos:
$y_\tau = \eos$.  We denote the length of $\y$ (including $\eos$) by $|\y|$.
We denote $\y_{<t} \coloneqq (y_1, \cdots, y_{t-1})$ for $t > 1$ and
we set $\y_{<1} \coloneqq \varnothing $. We denote the vocabulary augmented with
the \eos token as $\augvoc \coloneqq \cV \cup \{\eos\}$.  We use $\oplus$ to
indicate concatenation.  We denote random variables by capital letters, e.g.,
$Y$.  Given the finite set $\cY$, we define the space of probability
distributions \textit{with full support} over $\cY$ as $\cP(\cY)$ and the set of
functions $f \colon \cY \to \RR$ as $\cF(\cY)$.  Given a function $f \in
\cF(\cY)$, we define $\softargmax(f) \in \cP(\cY)$ by $\softargmax(f)(\y)
\coloneqq \frac{\exp f(\y)}{ \sum_{\y' \in \cY} \exp f(\y')}$.

\section{A unified perspective on EBMs and ARMs}
\label{sec:unified_perspective}

\subsection{Energy-based models (EBMs)}

\paragraph{Model definition.}

EBMs \citep{ackley1985learning,lecun2006tutorial} use a function $R \colon \cX
\times \cY \to \RR$,
measuring the ``affinity'' between an input $\x \in \cX$ (a prompt) 
and an output $\y \in \cY$ (a complete response).
We use the convention that higher value means higher affinity,
therefore $R$ can be interpreted as a negative energy.
EBMs are defined as
\begin{equation}
p^\ebm_R(\y|\x) 
\coloneqq \frac{\exp(R(\x, \y))}{\sum_{\y' \in \cY} \exp(R(\x, \y'))},
\label{eq:ebm_def}
\end{equation}
or more concisely,
$p^\ebm_R(\cdot|\x) = \mathrm{softargmax}(R(\x, \cdot))$.
We use the letter $R$ on purpose, as it will coincide with the notion of reward
function in RL.  The reward could be a learned model or a given function, such
as a verifier.

The associated sequence-level log-partition function is
\begin{equation}
A^\ebm_R(\x) 
\coloneqq \underset{\y \in \cY}{\lse} ~ R(\x, \y)
\coloneqq \log \sum_{\y \in \cY} \exp(R(\x, \y)).
\end{equation}
We then have
$p^\ebm_R(\y|\x) = \exp(R(\x, \y) - A^\ebm_R(\x))$.

Unfortunately, the log-partition costs $O(V^T)$ to compute
exactly.  This makes both learning EBMs and making inference with EBMs very
challenging.

\paragraph{EBMs as undirected graphical models.}

We can view EBMs as specifying a sequence-level
Gibbs (Boltzmann) distribution parameterized by a function $R$.
From a probabilistic graphical model perspective, EBMs are (conditional) 
Markov random fields, that is, undirected graphical models.

\paragraph{Validity of the distribution.}

Since all sequences $\y \in \cY$ have a maximal size $T$, and therefore $\cY$ is
finite, the denominator in \eqref{eq:ebm_def} is well-defined.  Exponentiation
and renormalization ensure that it is a valid probability distribution.

\textbf{Learning $R$ from data using Transformers.}
To learn $R$ from data, we must use a model that can cope with variable-length
sequences $\x$ and $\y$.  This can be achieved using a Transformer
\citep{vaswani2017attention}.  Because $R$ is sequence-level, the Transformer
can be non-causal (bidirectional).  A Transformer embeds an input sequence $\s =
\x \oplus \y$ of size $M$ into a latent space $\RR^D$, processing it through
multiple attention, feedforward and normalization layers, yielding $M$ output
vectors $(\h_1, \ldots, \h_M)$, where $\h_i \in \RR^D$. When using a non-causal
Transformer, each $\h_i \in \RR^D$ depends on the entire sequence $\s$. To
obtain a scalar-valued $R(\x, \y)$, one can for example use an affine layer on
top of a single representation $\bar \h$ of the sequence. The latter can be the
average of all $\h_i$, or simply the first vector $\h_1$, which mirrors the
\texttt{CLS} token in BERT encoders~\citep{devlin2019bert}. 

Learning $R$ from $(\x, \y)$ pairs is challenging due to the intractable
normalization constant, and has attracted a large body of work
\citep{song2021train,sander2025joint}.  To work around this difficulty, $R$ is
popularly learned from pairwise preferences $(\x, \y^{+}, \y^{-})$, where
$\y^{+}$ is preferred over $\y^{-}$ given $\x$
\citep{ziegler2019fine,stiennon2020learning,ouyang2022training} or from scored
pairs $(\x, \y, s)$, where $s$ is the affinity score (binary or real valued)
between $\x$ and $\y$, that $R(\x, \y)$ should approximate
\citep{cobbe2021training}.

\paragraph{Sampling.}

Given $R$, drawing \iid sequence samples
\begin{equation}
Y \sim p^\ebm_R(\cdot | \x)
\end{equation}
typically requires Markov-chain Monte-Carlo (MCMC) methods, such as Gibbs
sampling. However, such samplers are typically inherently sequential and
difficult to parallelize. Moreover, since they are never run to convergence in
practice, they do not generate truly \iid samples.

\paragraph{Optimal solution of MaxEnt RL as an EBM.}

It is well-known that an EBM can be seen as the solution of
\begin{equation}
p^\ebm_R = \argmax_{p \in \cP(\cY|\cX)} 
\EE_X \EE_{Y \sim p(\cdot | X)} R(X, Y) + H\big(p(\cdot|X)\big),
\end{equation}
where $H(\cdot)$ is the entropy of a distribution. More generally, RL-based
post-training of LLMs is commonly formulated as the KL-regularized problem 
\citep{ziegler2019fine,stiennon2020learning,ouyang2022training}
\begin{equation}
\argmax_{p \in \cP(\cY|\cX)} 
\EE_X \EE_{Y \sim p(\cdot | X)} R(X, Y) - \mathrm{KL}(p(\cdot|X), \pref(\cdot|X)),
\label{eq:reg_RL_objective}
\end{equation}
where $\pref \in \cP(\cY|\cX)$ is a reference ``anchor'' distribution.
The optimal solution $p^\star$ is then
\begin{equation}
p^\star(\y | \x) = \frac{\pref(\y|\x) \exp(R(\x, \y))}{
\sum_{\y' \in \cY}\pref(\y'|\x) \exp(R(\x, \y'))
}.
\end{equation}
Using \eqref{eq:ebm_def}, we can write $p^\star$ more concisely as
$p^\star = p^\ebm_{R + R_\mathrm{ref}}$,
where $R_\mathrm{ref}(\x, \y) \coloneqq \log \pref(\y | \x)$.
Therefore, the optimal solution of the KL-regularized RL objective
\eqref{eq:reg_RL_objective} is exactly an EBM, with the reward function $R +
R_\mathrm{ref}$.  Recently, several papers have replaced the KL divergence in
the regularization term by other $f$-divergences 
\citep{go2023aligning,wang2023beyond,roulet2025loss,sander2025joint}.

\subsection{Autoregressive models (ARMs)}

\paragraph{Model definition.}

Let us define the set of contexts $\cS \coloneqq \{\bos\} \times
\cup_{t=1}^{U+T-2} \voc^t$.  We can think of $\cS$ as the set of partial
(incomplete) sequences, including the initial prompt.  
Recall that we denoted $\augvoc \coloneqq \voc \cup \{\eos\}$
the augmented vocabulary used to encode finished responses.  
Given a pair $(\x, \y)$, where $\y = (y_1, \dots, y_\tau)$,
it will be convenient to define the context (state) at time $t$ by
\begin{equation}
\s_t \coloneqq
    \x \oplus \y_{<t} \\
 \in \cS(\x),
\label{eq:state_notation}
\end{equation}
where $\cS(\x) \subseteq \cS$ is the set of partial sequences prolonging the
prompt $\x$.
We assume that we
have at our disposal a function $q \colon \cS \times \augvoc$, for scoring the
``affinity'' between a context (state) $\s_t \in \cS$ and a next token $y_t \in
\augvoc$ (potentially \eos).  We can think of $q$ as a local scoring function,
while $R$ used by EBMs is a global scoring function.
ARMs \citep{bengio2003neural} use the chain rule of probability to define
the probability of a sequence $\y$ given $\x$ as
\begin{equation}
p^\arm_q(\y|\x) \coloneqq
\prod_{t=1}^{|\y|} \pi_q(y_t|\underbrace{\x \oplus \y_{<t}}_{\s_t}),
\end{equation}
where
\begin{equation}
\pi_q(y_t|\s_t) \coloneqq 
\frac{\exp(q(\s_t, y_t))}{\sum_{j \in \augvoc} \exp(q(\s_t, j))}.
\end{equation}
The latter can be rewritten more concisely as
$\pi_q(\cdot|\s_t) = \mathrm{softargmax}(q(\s_t, \cdot))$.
The associated log-partition is
\begin{equation}
V_q(\s_t)
\coloneqq \underset{j \in \augvoc}{\lse} ~ q(\s_t, j)
\coloneqq \log \sum_{j \in \augvoc} \exp(q(\s_t, j)).
\end{equation}
We then have
$\pi_q(y_t|\s_t) = \exp(q(\s_t, y_t) - V_q(\s_t))$.

\paragraph{ARMs as directed graphical models.}

From a probabilistic graphical model perspective, ARMs are Bayesian networks, 
that is, directed graphical models.
More specifically, we can view ARMs as specifying a time-inhomogeneous
infinite-order Markov chain (assuming unlimited context window size)
with discrete states $Y_t$.

\paragraph{Validity of the distribution.}

For $p^\arm_q(\cdot|\x)$ to define a valid probability distribution over all
possible sequences of size at most $T$, we need to enforce that sequences
cannot be of size more than $T$. 
This can be achieved by fixing 
\begin{equation}
q(\s_T, y_T) \coloneqq
\begin{cases}
0 &\mbox{if} ~ y_T = \eos \\
-\infty &\mbox{if} ~ y_T \neq \eos
\end{cases}.
\label{eq:q_last_token}
\end{equation}
This ensures that $\pi_q(\eos|\s_T) = 1$, which in turn ensures that the ARM
defines a valid probability distribution over all possible sequences of size at
most $T$.  See Appendix \ref{sec:variable_length_handling} for more details and
a proof.

\paragraph{Analogy with EBMs.}

We now rewrite the equations in order to build an analogy with EBMs.
We define the ``log-partition'' \textbf{along} the
sequence (path) $\y$ given $\x$ as
\begin{equation}
A^\arm_q(\x, \y) 
\coloneqq \sum_{t=1}^{|\y|} V_q(\x \oplus \y_{<t})
= \sum_{t=1}^{|\y|} \underset{y'_t \in \augvoc}{\lse} ~ q(\x \oplus \y_{<t}, y'_t).
\end{equation}
Mirroring the EBM notation, we then have
\begin{equation}
p^\arm_q(\y|\x) = \exp(\Qsub{q}(\x, \y) - A^\arm(\x, \y)),
\end{equation}
where
\begin{equation}
\Qsub{q}(\x, \y) \coloneqq \sum_{t=1}^{|\y|} q(\x \oplus \y_{<t}, y_t).
\end{equation}

\paragraph{Sampling.}

To draw an \iid sequence $Y \sim p^\arm_q(\cdot|\x)$, 
we can use autoregressive (\aka ancestral) sampling,
\begin{equation}
Y_t \sim \pi_q(\cdot | S_t),
\end{equation}
where $S_1 \coloneqq \x$ and $S_t \coloneqq \x \oplus Y_{<t}$ for $t > 1$.
The sampling is repeated until an \eos token is emitted. 
\paragraph{Learning $q$ from data using Transformers.}

To learn $q$ from data, we must use a model that can score a next token $y_t$
using only the past context $\s_t$, which has variable length.  This can be
achieved using a causal Transformer $\cT$, which uses an attention mask to
ensure that each token can only attend to previous tokens.  The causal
Transformer outputs a vector of logits $\h_t = \cT(\s_t) \in \RR^{|\cA|}$.
Intuitively, $\cT(\s_t)[y_t]$ can be interpreted as the score of token $y_t \in
\cA$ given the context $\s_t$.  We can therefore define $q(\s_t, y_t) \coloneqq
\cT(\s_t)[y_t].$

\paragraph{MaxEnt RL as distilling an EBM into an ARM.}

We saw that the optimal solution of the regularized RL problem
\eqref{eq:reg_RL_objective} is exactly an EBM.  This creates a paradox: if we
know the optimal solution, why cannot we use the EBM directly at inference time
and why do we even need RL-based training? Unfortunately, as we discussed,
sampling from an EBM
in combinatorially-large spaces is typically intractable, often requiring
Markov-chain Monte-Carlo (MCMC) methods. Therefore, \eqref{eq:reg_RL_objective}
is often reformulated as
\begin{equation}
\begin{aligned}
&\argmax_{q \in \cF(\cS \times \augvoc)} 
\EE_X \EE_Y R(X, Y) - \mathrm{KL}(p^\arm_q(\cdot|X), \pref(\cdot|X)) \\
=& \argmin \EE_X \mathrm{KL}(p^\arm_q(\cdot|X), p^\ebm_{R + R_{\mathrm{ref}}}(\cdot|X))
\end{aligned}
\end{equation}
where $Y$ above is distributed according to $p^\arm_q(\cdot|X)$.

Intuitively, instead of performing maximization in the space $\cP(\cY|\cX)$ of
all possible distributions, we perform maximization in the space of ARMs
parameterized by a function $q \in \cF(\cS \times \cA)$.  
Thus, we can see RL-based language model post-training
as approximating the EBM  $p^\ebm_{R + R_\mathrm{ref}}(\cdot | \x)$ with the ARM
$p^\arm_q(\cdot|\x)$, or put
differently, as distilling this EBM into this ARM.

\begin{figure}[t]
\centering
\includegraphics[width=0.7\linewidth]{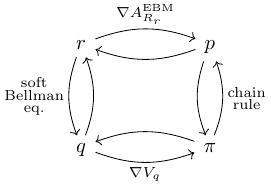}
\caption{Summary of mappings discussed in this paper.}
\label{fig:diagram}
\end{figure}

\section{From the chain rule to Bellman equations}

\subsection{Bijection between distributions}
\label{sec:bijection_distributions}

The chain rule of probability factorizes a joint probability
distribution into a product of conditional probabilities.
In this section, we show how it can be used to derive a procedure for
converting any sequence distribution $p \in \cP(\cY|\cX)$ to a 
next-token distribution $\pi \in \cP(\augvoc|\cS)$, and vice-versa.

\paragraph{From sequence to next-token distributions.}

For any $\x \in \cX$ and any $\y \in \cY$ such that $|\y| = \tau$, with $1 \le
\tau \le T$, the procedure works by successively marginalizing and conditioning
in backward order.  First, starting from $p(\y_{\le \tau}|\x) \coloneqq p(\y |
\x)$, we define from $t=\tau-1$ to $t=1$,
\begin{equation}
p(\y_{\le t}|\x) \coloneqq \sum_{y_{t+1} \in \augvoc} 
p(\y_{\le t}, y_{t+1}|\x).
\end{equation}
Second, 
we define from $t=\tau$ to $t=2$
\begin{equation}
\pi(y_t|\x, \y_{<t}) \coloneqq \frac{p(\y_{\le t}|\x)}{p(\y_{\le t-1}|\x)},
\end{equation}
ending with $\pi(y_1|\x) \coloneqq p(y_1|\x)$.

\paragraph{From next-token to sequence distributions.}

Conversely, from any next-token distribution $\pi \in
\cP(\augvoc|\cS)$, we can reconstruct a sequence distribution $p
\in \cP(\cY|\cX)$ by
\begin{equation}\label{eq:chain_rule_of_proba}
p(\y|\x) = \prod_{t=1}^{|\y|} \pi(y_t | \x, \y_{<t}). 
\end{equation}
Indeed, by substituting the definition of $\pi$, we observe that the
product forms a telescoping sequence that exactly recovers the joint
distribution:
$
\prod_{t=1}^{|\y|} \pi(y_t | \x, \y_{<t}) 
= \frac{p(\y_{\le 1}|\x)}{1} \times
\frac{p(\y_{\le 2}|\x)}{p(\y_{\le 1}|\x)} \times \dots \times \frac{p(\y_{\le
\tau}|\x)}{p(\y_{\le \tau-1}|\x)}
= p(\y|\x)$.

\paragraph{The chain rule of probability as a bijection.}

Since there is a unique correspondence between $p$ and $\pi$, the procedure
defines a bijection between $\cP(\cY|\cX)$ and $\cP(\cA|\cS)$.

\subsection{Bijection in function space}
\label{sec:bijection}

We now instantiate the bijection for EBMs and ARMs, which
can be seen as using sequence and next-token Gibbs (Boltzmann)
distributions, respectively. In this section, we assume without loss of
generality that $R$ decomposes as
\begin{equation}
\Rsub{r}(\x, \y) \coloneqq \sum_{t=1}^{|\y|} r(\underbrace{\x \oplus \y_{<t}}_{\s_t}, y_t),
\label{eq:R_decomposition}
\end{equation}
where $r \colon \cS \times \augvoc \to \RR$ is the immediate (next-token) reward.
Indeed, if $R$ does not naturally decompose as above,
we can always assign all the non-zero reward in the last step.

We now show how to transform 
$r \colon \cS \times \augvoc \to \RR$
used by an EBM into
$q \colon \cS \times \augvoc \to \RR$
used by an ARM, and vice-versa.

\paragraph{From EBM to ARM.} 

Suppose $r \colon \cS \times \augvoc \to \RR$ is given. 
We define the mapping $q = \cM(r)$ as
\begin{equation}
q(\s_t, y_t) \coloneqq
\begin{cases}
r(\s_t, y_t) &\mbox{if} ~ y_t = \eos \\
r(\s_t, y_t) + V_q(\s_t \oplus y_t) &\mbox{if} ~ y_t \neq \eos
\end{cases},
\label{eq:q_from_r}
\end{equation}
where $\s_t \in \cS$ and $y_t \in \augvoc$.
Recall that
\begin{equation}
V_q(\s_t \oplus y_t) = \underset{y_{t+1} \in \augvoc}{\lse} ~ q(\s_t \oplus y_t, y_{t+1}). 
\end{equation}
We can see that $q(\s_t, \cdot)$ depends on $q(\s_{t+1}, \cdot)$, creating a recursive dependency.
Computing $q$ from $r$ must thus be done sequentially from $t=\tau$ to $t=1$ for all possible states.

\paragraph{From ARM to EBM.} 

Suppose $q \colon \cS \times \augvoc \to \RR$ is now given. 
We define the inverse mapping $r = \cM^{-1}(q)$ as
\begin{equation}
r(\s_t, y_t) \coloneqq
\begin{cases}
q(\s_t, y_t) &\mbox{if} ~ y_t = \eos \\
q(\s_t, y_t) - V_q(\s_t \oplus y_t) &\mbox{if} ~ y_t \neq \eos
\end{cases}.
\label{eq:r_from_q}
\end{equation}
As there is no longer a recursive dependency,
computing $r$ from $q$ can be done in parallel.

\paragraph{Bijection.} 

Having defined the mapping $\cM$ and its inverse $\cM^{-1}$,
we can now formally state our result.

\begin{proposition}[Bijection in function space]\label{prop:bijection}
The mapping $q = \cM(r)$ is bijective and
for all $\x \in \cX$ and $\y \in \cY$, we have
\begin{align}
    p^\ebm_{\Rsub{r}}(\y | \x) &= p^\arm_q(\y | \x) \\
    A^\ebm_{\Rsub{r}}(\x) &= V_q(\x).
\end{align}
\end{proposition}
A proof is given in Appendix \ref{proof:bijection}.
This shows that an EBM can be transformed into an
ARM if $r$ is ``corrected'' by adding the log-partition $V_q$ of the next
states.  In MaxEnt RL terminology (see Section \ref{sec:maxent_rl_perspective}
for more details), $V_q$ is known as the soft value function and $V_q(\s_t)$
captures the future (soft) value of being in state $\s_t$. 
Accordingly, since an ARM has the capacity to learn an EBM,
which is a globally normalized distribution,
an ARM inherently has the ability to look ahead, provided that we can
learn $q$ appropriately.

\paragraph{Computational cost.}

The cost of converting an ARM to an EBM is $O(VT)$,
linear in the sequence length. 
On the other hand, the cost of converting an EBM into an ARM is $O(V^T)$, exponential in the
sequence length. Explicitly converting an EBM into an ARM is therefore intractable.
However, as we shall discuss in Section \ref{sec:minima}, when learning from $(\x, \y)$ pairs,
such an explicit conversion is not needed, as the optimal ARM is \textit{implicitly} an EBM!

\subsection{Maximum-entropy RL perspective}
\label{sec:maxent_rl_perspective}

\paragraph{Autoregressive models as MDPs.}

Let $\cS$ be the set of states and $\cA$ be the set of actions.
ARMs can be seen as a special case of Markov decision
processes (MDPs) where
\begin{itemize}[topsep=0pt,itemsep=3pt,parsep=3pt,leftmargin=16pt]
\item states are contexts: $\s_t \coloneqq \x \oplus \y_{<t} \in \cS$;
\item actions are tokens: $y_t \in \augvoc$;
\item next states append the last token:
    $\s_{t+1} \coloneqq \s_t \oplus y_t$;
\item horizon is finite and there is no decay factor ($\gamma = 1$);
\item $r$ is the immediate reward function.
\end{itemize}
From this perspective, an ARM is a particular instance
of a MDP, where the transition kernel $\mu(\cdot|\s_t, y_t)$ is deterministic
with $\mu(\s_t \oplus y_t|\s_t, y_t) = 1$.  These particularly
simple state transition dynamics imply that each
state is never visited more than once. 
This perspective should not be confused with the (combinatorial) contextual
bandit perspective, where $\x$ corresponds to the single state, $\y$ corresponds
to an action and $R$ is the sequence-level reward function.

\textbf{Soft Bellman equation.} In MaxEnt RL, we seek the policy
\begin{equation}
\pi^\star = \argmax_{\pi \in \cP(\cA|\cS)}    
\EE \sum_{t=1}^\infty \gamma^t (r(S_t, A_t) - \log \pi(A_t | S_t)).
\label{eq:policy_gradient_objective}
\end{equation}
The expectation is over trajectories of state-action pairs 
starting from $S_1 \sim p_\cX$ and following
\begin{align}
A_t &\sim \pi(\cdot | S_t) \\
S_{t+1} &\sim \mu(\cdot | S_t, A_t).
\end{align}
It is well-known in this literature (see Section \ref{sec:related_work} for a
detailed review) that an optimal solution is
\begin{equation}
\pi^\star(a|\s) = \exp(q^\star(\s, a) - V_{q^\star}(\s))
\end{equation}
where $q^\star$ is a solution of the fixed point equation
\begin{equation}
q(\s, a) = r(\s, a) + \gamma \EE_{S \sim \mu(\cdot | \s, a)} V_{q}(S).
\label{eq:soft_q_learning_objective}
\end{equation}

\paragraph{Existence and uniqueness of the fixed point.}

In the general RL setting (infinite horizon, cyclic MDPs), the existence and
uniqueness of the fixed point $q^\star$ rely on the Banach fixed point theorem.
The standard condition 
is simply $\gamma < 1$, which guarantees that the soft Bellman operator
$(\mathcal{B}_r q)(\s, a) \coloneqq r(\s, a) + \gamma \EE_{S \sim \mu(\cdot |
\s, a)} V_{q}(S)$ is a contractive mapping.
Typically, $q^\star$ does not enjoy an explicit formula and
finding $q^\star$ requires some form of fixed point iterations.
In contrast, in our setting (finite horizon, acyclic MDP), 
$q^\star = \cM(r)$ is the explicit formula of the fixed point solution, which
guarantees both existence and uniqueness (invertibility).
Moreover, we constructed the bijection through the chain rule of probability,
unraveling a deep connection with the soft Bellman equation
in our particular setting.

\subsection{Entropy-regularized DP on a DAG perspective}
\label{sec:dp_dag}

We can also view our bijection as regularized dynamic programming (DP) on a
directed acyclic graph (DAG).  In this perspective, we define a DAG where
\begin{itemize}[topsep=0pt,itemsep=3pt,parsep=3pt,leftmargin=16pt]
\item there is one start (root) node corresponding to $\s_1 = \x$;
\item there is one end node;
\item intermediate nodes are contexts $\s_t \coloneqq \x \oplus \y_{<t}$;
\item edge weights indicate the value $r(\s_t, y_t)$ of appending $y_t$ to
    context $\s_t$ (higher is better);
\item nodes such that $\s_t \coloneqq \x \oplus (y_1, \dots, \eos)$ have a
    single out-going edge of weight $0$, connected to the end node.
\end{itemize} 
Computing
$\max_{\y \in \cY} \sum_{t=1}^{|\y|} r(\x \oplus \y_{<t}, y_t)$ 
then amounts to finding the path in the DAG of maximum value
and computing the log-partition
$A^\ebm_{\Rsub{r}}(\x) 
= \underset{\y \in \cY}{\lse} ~ \sum_{t=1}^{|\y|} r(\x \oplus \y_{<t}, y_t)$
amounts to finding the path of {\em soft} maximum value.

\paragraph{Top-down vs.\ bottom-up DP.}

To compute the soft-maximum 
$A^\ebm_{\Rsub{r}}(\x) = V_q(\x)$
for $q = \cM(r)$ by DP, two approaches are possible.
In \textit{top-down} DP, we start from $V_q(\x)$, the quantity we wish to
compute, and recurse. This corresponds to traversing the DAG forward in time.
In \textit{bottom-up} DP (tabulation), we start from the base cases and make our
way up to $V_q(\x)$.  This corresponds to traversing the DAG backward in time.
As shown in \citep{mensch2018differentiable}, the associativity of the
log-sum-exp and of the distributivity of $+$ over the log-sum-exp guarantee the
optimality of dynamic programming. Indeed, we have (fixing $|\y| = T$ to
simplify the notation)
\begin{small}
\begin{align}
A^\ebm_{\Rsub{r}}(\x)
&= \underset{\y \in \cY}{\lse} ~ \sum_{t=1}^T r(\x \oplus \y_{<t}, y_t) \\
&= \underset{y_1 \in \cV}{\lse} \dots \underset{y_T \in \cV}{\lse} ~
\sum_{t=1}^T r(\x \oplus \y_{<t}, y_t) \\
&= \underset{y_1 \in \cV}{\lse} ~ \biggl[ r(\x, y_1)
+ \dots + \underset{y_T \in \cV}{\lse} ~ \Bigl[ r(\x \oplus \y_{<T}, y_T) \Bigr] \biggr].
\end{align}
\end{small}
It is well-known that the gradient of the log-partition exactly coincides
with marginal probabilities \citep{wainwright2008graphical}.
In Appendix \ref{sec:backpropagation},
we show that the gradient \wrt $r$ of
$A^\ebm_{\Rsub{r}}(\x) = V_q(\x)$ for $q = \cM(r)$ indeed exactly coincides with the
marginal probabilities of response prefixes.

\section{Theoretical analysis}

\subsection{Learning EBMs and ARMs from input-output pairs}
\label{sec:minima}

In this section, 
building upon our bijection in Section \ref{sec:bijection},
we state equivalence results between learning EBMs and learning ARMs from
$(\x, \y)$ pairs.

\paragraph{Negative log-likelihood of EBMs.}

The negative log-likelihood of $\y$ given $\x$ according to $p^\ebm_R$ is
\begin{equation}\label{eq:lebm}
\ell^\ebm_R(\x, \y) 
\coloneqq -\log p^\ebm_R(\y|\x) 
= A^\ebm_R(\x) - R(\x, \y).
\end{equation}
The loss is convex in $R$ and satisfies the property
\begin{equation}
\ell^\ebm_R(\x, \y) = 0 \iff p^\ebm_R(\y|\x) = 1.
\end{equation}

\paragraph{Negative log-likelihood of ARMs.}

The negative log-likelihood of $\y$ given $\x$ according to
$p^\arm_q$ is
\begin{align}
\ell^\arm_q(\x, \y)
&\coloneqq - \log p^\arm_q(\y|\x) \\
&= A^\arm(\x, \y) - \Qsub{q}(\x, \y).
\end{align}
The loss is convex in $q$ and satisfies the property
\begin{equation}
\ell^\arm_q(\x, \y) = 0 \iff p^\arm_q(\y|\x) = 1.
\end{equation}

Unlike $A^\ebm$, $A^\arm$ is a function of $\x$ and $\y$, not just $\x$.
Therefore, contrary to EBMs, the negative log-likelihood of an ARM is using the
ground-truth sequence $\y$ as a path. This is known as \textbf{teacher
forcing}. Despite what the name may suggest, this behavior arises naturally as a
consequence of using negative log-likelihood on ARMs.

\paragraph{Equivalence of minima.}

Let us define the expected risks
\begin{align}
\cL^\arm(q) &\coloneqq \EE_{(X,Y) \sim \rho} \left[\ell^\arm_q(X, Y)\right] 
\label{eq:expected_risk_arm} \\
\cL^\ebm(R) &\coloneqq \EE_{(X,Y) \sim \rho} \left[\ell^\ebm_R(X, Y)\right]. 
\label{eq:expected_risk_ebm}
\end{align}
We then have the following equivalence, which is a straightforward consequence
of our bijection in Section \ref{sec:bijection}.
\begin{proposition}[Equivalence of minima]\label{prop:same_minima}

For any joint distribution $\rho$ over $\cX \times \cY$, we have
\begin{align}
\min_{q \in \cF(\cS \times \augvoc)} \cL^\arm(q)
&= 
\min_{r \in \cF(\cS \times \augvoc)} 
\cL^\ebm(\Rsub{r}) \\
&= 
\min_{R \in \cF(\cX \times\cY)} 
\cL^\ebm(R)
\end{align}
and, with $q^\star = \cM(r^\star)$, we have
\begin{small}
\begin{align}
q^\star \in \argmin_{q \in \cF(\cS \times \augvoc)}
\cL^\arm(q) 
&\iff
r^\star \in \argmin_{r \in \cF(\cS \times \augvoc)} 
\cL^\ebm(\Rsub{r}) \\
&\implies
\Rsub{r^\star} \in \argmin_{R \in \cF(\cX \times \cY)} 
\cL^\ebm(R).
\end{align}
\end{small}
\end{proposition}
See Appendix \ref{proof:same_minima} for a proof.
Proposition \ref{prop:same_minima} shows that, when the goal is to fit a model
to observed input-output pairs $(\x, \y)$ by negative log-likelihood minimization, EBMs and
ARMs are equivalently powerful, provided that we perform minimization in
function space, which means that the functions are able to fit the data
perfectly.

\paragraph{Optimality of teacher forcing.}

Learning the reward $R$ from data corresponds to an \textbf{inverse RL} setting
\citep{ziebart2008maximum}.  From a duality perspective \citep{sander2025joint},
the optimal primal variable $R^\star$ is linked to the optimal dual variable
$p^\star$ through, for all $\x \in \cX$,
\begin{equation}
p^\star(\cdot | \x) 
= \mathrm{softargmax}(R^\star(\x, \cdot)) 
= p^\ebm_{R^\star}(\cdot | \x). 
\end{equation}
Unfortunately, recovering $p^\star$ from $R^\star$ is difficult, due to the
intractable normalization constant.  However, given $q^\star$ (the optimal
solution via teacher forcing), $p^\star$ can be constructed from $\pi_{q^\star}$
using the chain rule of probability in \eqref{eq:chain_rule_of_proba},
\begin{equation}\label{eq:p_star_from_q_star}
p^\star(\y|\x) = \prod_{t=1}^{|\y|} \pi_{q^\star}(y_t | \x, \y_{<t}). 
\end{equation}
This demonstrates that teacher forcing is optimal when optimizing in function
space. We stress that the mapping $\cM$ is completely implicit
in this case (we do not need to explicitly compute it).
We summarize the mappings in Figure \ref{fig:diagram}.

\subsection{Distilling EBMs into ARMs}

We saw in Section \ref{sec:unified_perspective} that KL-regularized RL can be
seen as distilling EBMs into ARMs.  This motivates the need for approximation
guarantees of EBMs by ARMs.

\paragraph{Function space.}

Stated differently, Proposition \ref{prop:bijection} shows that the exact
mapping $q = \cM(r)$ satisfies
for all $\x \in \cX$ 
\begin{equation}
\KL(p^\arm_q(\cdot|\x), p^\ebm_{\Rsub{r}}(\cdot|\x)) = 0.
\end{equation}
Such a result holds in function space.  This is the so-called tabular setting in
RL, where the functions $q \colon \cS \times \augvoc \to \RR$ and $r \colon \cS
\times \augvoc \to \RR$ are represented as (gigantic) arrays, exploiting the
fact that the space of contexts (states) $\cS$ and the space of next tokens
$\augvoc$ are finite. 

\paragraph{Function approximation setting.}

In practice, $q$ and $r$ are implemented using parameterized functions.  In this
setting, there may no longer be an exact conversion between EBMs and ARMs, but we
can bound their KL divergence.
\begin{proposition}[KL bound]
For all
$r \colon \cS \times \augvoc \to \RR$,
$q \colon \cS \times \augvoc \to \RR$
and
$\x \in \cX$, we have
\begin{equation}
\KL(p^\arm_q(\cdot|\x), p^\ebm_{\Rsub{r}}(\cdot|\x))
\le 2T \max_{\substack{\s \in \cS(\x)\\ y \in \augvoc}} |q^\star(\s, y) - q(\s, y)|,
\end{equation}
where
$q^\star \coloneqq \cM(r)$.
\label{prop:kl_bound}
\end{proposition}
A proof is given in Appendix \ref{proof:kl_bound}.
Note that the KL divergence is not symmetric but the same bound 
also applies for
$\KL(p^\ebm_{\Rsub{r}}(\cdot|\x), p^\arm_q(\cdot|\x))$.

\paragraph{Approximation using a Transformer.}

For a fixed EBM parameterized by $\Rsub{r}$, let $q^\star = \cM(r)$.  We now show that
$q^\star$ can be approximated by a causal Transformer model by leveraging a
powerful approximation result of causal mappings by causal Transformers
\citep[Theorem 2]{furuya2024transformers}.  More precisely, for all $\varepsilon
> 0$,  there exists a causal Transformer $\mathcal{T}: \cS \to \RR^{|\augvoc|}$ such
that, for all $\x \in \cX$, 
\begin{equation}
\max_{\substack{\s \in \cS(\x), y \in \augvoc}}|q^\star(\s,y)- \mathcal{T}(\s)[y]|\leq \varepsilon.
\end{equation}
Together with Proposition \ref{prop:kl_bound}, this gives for all $\x \in \cX$,
\begin{equation}
\KL(p^\arm_{\mathcal{T}(.)[.]}(\cdot|\x), p^\ebm_{\Rsub{r}}(\cdot|\x))
\le 2T \varepsilon.
\end{equation}
However, this result requires the embedding dimension of $\cT$ to
scale with the vocabulary size $|\mathcal{V}|$ \citep{furuya2024transformers}.

\section{Numerical validation}

In this section, we validate our theory using synthetic tiny language models. We
focus on small vocabulary $V$ and small maximum length $T$ , enabling exact
computation of the EBM's log-partition, without resorting to approximate
algorithms.  We minimize the expected risks in \eqref{eq:expected_risk_arm} and
\eqref{eq:expected_risk_ebm}, using a non-causal Transformer for the EBM and
a causal Transformer for the ARM. 
The minimum risk is defined as $\cL^\star \coloneqq -\EE_{(X,Y) \sim \rho} \log
\rho(X, Y)$. All experimental details are included in
Appendix~\ref{app:exp_details}.

\begin{figure}[t]
    \centering
    \includegraphics[width=\linewidth]{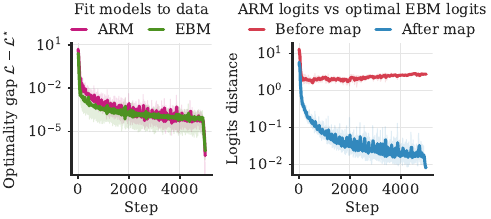}
    \caption{Empirical validation of Proposition \ref{prop:same_minima}.
    \textbf{Left}: Minimizing the expected risk of an ARM and an EBM
parameterized by causal and non-causal Transformers, respectively.
\textbf{Right}: $L_\infty$ distance between the logits of the trained ARM and
the logits of the \textbf{optimal} EBM, before and after applying the mapping
$\cM$.}
    \label{fig:ebm_arm_bij_illus}
\end{figure}

In Figure~\ref{fig:ebm_arm_bij_illus} (left), we plot the optimality gap $\cL -
\cL^\star$ of both the EBM and the ARM as a function of optimization steps.  Our
results confirm that the EBM and ARM converge to the same minima, as predicted
by Proposition~\ref{prop:same_minima}.  Perhaps more surprisingly, our results
also show that the loss curves of both models are very close, during the entire
training course.  These findings are confirmed across several model sizes, and
sequence-length / vocabulary ratios (Appendix~\ref{app:exp_details}).

In Figure~\ref{fig:ebm_arm_bij_illus} (right), we show the $L_\infty$ distance
between the logits of the trained ARM and the logits of the optimal EBM, before
and after applying the mapping $\cM$.  Our results therefore confirm that the
mapping $\cM$ can be used to explicitly convert the logits of the EBM and ARM.
However, we emphasize that explicit logit conversion is not needed, since we can
also factorize the EBM distribution using \eqref{eq:p_star_from_q_star}.

\section{Related work}
\label{sec:related_work}

\paragraph{Maximum-entropy RL.}

Maximum entropy was used for inverse RL, where the goal is to recover the reward
function from offline trajectories \citep{ziebart2008maximum}.  In the forward
setting, a precursor is dynamic policy programming \citep{azar2012dynamic}.
Many papers subsequently formally established the equivalence between
policy-based methods, that seek to solve \eqref{eq:policy_gradient_objective}
\wrt $\pi$, and value-based methods, that seek to satisfy
\eqref{eq:soft_q_learning_objective} \wrt $q$
\citep{o2016pgq,haarnoja2017reinforcement,nachum2017bridging,schulman2017equivalence}.
The equivalence was revisited in continuous action space using convex duality
\citep{richemond2017short}.
Entropy-regularized RL was further studied \citep{neu2017unified} and extended
to other strongly-convex regularizers \citep{geist2019theory}.
Several papers proposed one-step and multi-step consistency loss functions based
on applying the squared loss to the optimality condition
\eqref{eq:soft_q_learning_objective}
\citep{nachum2017bridging,schulman2017equivalence,clavier2025shiq}.
Several papers proposed soft actor-critic algorithms
\citep{o2016pgq,haarnoja2017reinforcement,haarnoja2018soft} or joint learning
of the policy and value function \citep{richemond2024offline}.

\paragraph{RL as probabilistic inference.}

Maxent RL can be viewed from a probabilistic inference perspective, using
undirected graphical models
\citep{toussaint2009probabilistic,ziebart2010modeling}, defining EBMs (\aka
conditional random fields), or directed graphical models, mirroring hidden
Markov models  \citep{attias2003planning,levine2018reinforcement}.  Notably,
this connection has been leveraged to solve inference problems using MCTS and
learned soft value functions \citep{buesing2020approximate}.  Conversely,
probabilistic inference tools can be leveraged for maxent RL
\citep{korbak2022rl}, such as twisted sequential Monte Carlo
\citep{zhao2024probabilistic}.
We revisit the probabilistic inference perspective in our setting
in Appendix \ref{sec:proba_inference_perspective}.

\paragraph{Energy-based models.}

The standard approach for learning an EBM from $(\x, \y)$ pairs by stochastic
gradient descent requires sampling from the EBM \citep{song2021train}, which
usually involves MCMC algorithms such as Langevin when $\cY$ is continuous or
Gibbs sampling when $\cY$ is discrete. Since that approach learns the function $R$ (the reward in inverse RL; see Section \ref{sec:minima}), and not a
generative model $\pi$, MCMC is needed for generating samples from the EBM
parameterized by the learned $R$.  A min-min formulation was proposed to jointly
learn $R$ and the log-partition $A$ \citep{sander2025joint}. While this bypasses
the need for MCMC at train time, it still requires it for sampling from the EBM.
We emphasize however that if we do not need to sample but just to predict the
mode of the EBM (the most likely output $\y$ given $\x$), then we only need to
maximize $R$, which does not involve the log-partition and is tractable for some
sets $\cY$.  On the other hand, min-max formulations can be used to jointly
learn a discriminator (reward) $R$ and a generator (policy) $\pi$ from $(\x,
\y)$ pairs \citep{ho2016generative}. Such approach does not require MCMC if the
generator easily generates samples by design, as is the case with ARMs.

\paragraph{GFlowNets.}

GFlowNets \citep{bengio2021flow,bengio2023gflownet} can be used to perform
amortized probabilistic inference \citep{hu2023amortizing}. Flow consistency equations play the same role
as soft Bellman equations, enabling the design of consistency loss functions
\citep{malkin2022trajectory}.  In fact, recent work has established that
GFlowNets can be cast as MaxEnt RL, even in the general DAG setting
\citep{tiapkin2024generative}.

\paragraph{Limitations of teacher forcing.} 

Recent work argued that the next-token prediction objective (teacher forcing) is
structurally ``myopic'' and fails to backpropagate future value information
effectively, leading to a failure in planning \citep{bachmann2024pitfalls}.  In
contrast, Proposition \ref{prop:same_minima} shows that the global minimum of
the teacher forcing objective (in function space) \textit{is} an EBM, which
inherently performs lookahead via the partition function $V_q$. Our work
therefore suggests that the limitation of teacher forcing is not in the
objective itself but in the difficulty of reaching its minimum, due to
optimization or approximation error.  This aligns with prior work showing that
planning failures stem from overfitting to spurious correlations under weak
supervision, rather than inherent structural incapacity
\citep{frydenlund2024mystery}.
In complementary work, behavior cloning, the counterpart of teacher forcing in RL, was shown to achieve horizon-independent sample complexity
\citep{foster2024behavior}.

\section{Conclusion}

\paragraph{Significance of this work.}

EBMs and ARMs are usually thought as two very distinct classes of models.  EBMs
are globally consistent but are difficult to train and sample from, while ARMs
are easy to train and sample from but are only backward looking.  Our paper
shows that EBMs and ARMs are in fact intimately related.  Our first contribution
is to establish a bijective mapping $\cM$ between EBMs and ARMs in function
space.  While this mapping is a special case of soft Bellman equation in general
MDPs (Section \ref{sec:maxent_rl_perspective}) and of entropy-regularized
dynamic programming in a DAG (Section \ref{sec:dp_dag}), we arrive at it from a
very different angle, that of the chain rule of probability (Section
\ref{sec:bijection_distributions} and Section \ref{sec:bijection}).  When
learning from $(\x, \y)$ pairs, Proposition \ref{prop:same_minima} shows that,
in function space, EBMs and ARMs are equivalently powerful, establishing the
optimality of teacher forcing in this case.  When distilling an EBM teacher into
an ARM student, 
Proposition
\ref{prop:kl_bound} shows that the KL between the EBM and the ARM is bounded by
the $L_\infty$ norm of the bijective mapping $\cM$.
Our results, validated through numerical experiments,
provide some justification for the next-token prediction
paradigm and for teacher forcing.

\paragraph{Limitations.}

When learning from $(\x, \y)$ pairs, Proposition \ref{prop:same_minima}
establishes that ARMs and EBMs are equally powerful when optimizing for $R$ and
$q$ in function space. However, in practice, $R$ can be a
non-causal Transformer while $q$ must be causal.  In addition, Proposition
\ref{prop:bijection} establishes that, in order for ARMs and EBMs to be
equivalent, $q$ must take the form $q(\s, y) = r(\s, y) + V_q(\s \oplus y)$,
where $V_q$ represents the (soft) optimal value of all future continuations. The
function $q$ is therefore burdened with a dual task: it must model the immediate
local score $r$, and it must learn to implicitly compute a potentially complex
future-looking value function $V_q$ (marginalizing over all possible futures)
using only a backward-looking architecture.  
Proposition \ref{prop:same_minima} does not account for optimization or
approximation error.
Formally comparing the complexity
of learning $R$ versus that of learning $q$ is a promising future direction.
Another important direction is to study the impact of latent variables (thinking
traces) in enhancing the expressivity of ARMs.

\paragraph{Bridging communities.}

This paper builds bridges between ARMs, EBMs and Maxent RL.
We hope that our paper contributes to shedding additional light on how many
works and perspectives relate to each other.

\section*{Impact statement}

This paper studies the connection between energy-based models and autoregressive models to justify the lookahead capabilities of next-token prediction. 
We do not foresee any specific ethical or societal implications arising directly from this work.

\bibliography{main}
\bibliographystyle{icml2026}

\newpage
\appendix
\onecolumn

\section{Supplementary materials}
\label{sec:supp_material}

Table of contents:
\begin{itemize}[topsep=0pt,itemsep=3pt,parsep=3pt,leftmargin=16pt]

\item Handling variable-length sequences 
\begin{itemize}
    \item In ARMs (Section \ref{sec:variable_length_handling}).
    \item In EBMs (Section \ref{sec:variable_length_handling_ebms}).
\end{itemize}
\item Chain rule 
\begin{itemize}
    \item Chain rule of entropy (Section \ref{sec:chain_rule_of_entropy}).
    \item Chain rule of KL (Section \ref{sec:chain_rule_of_kl}).
\end{itemize}
\item Bijection
\begin{itemize}
\item Bijection with a reference measure 
    (Section \ref{sec:bijection_with_ref_measure}).
\item Chain rule of probability perspective 
    (Section \ref{sec:bijection_chain_rule_perspective}).
\item Variational perspective 
    (Section \ref{sec:bijection_variational_perspective}).
\item Adversarial and variational inference perspectives 
    (Section \ref{sec:bijection_adversarial_perspective}).
\item Probabilistic inference perspective 
    (Section \ref{sec:proba_inference_perspective}).
\end{itemize}
\item Gradients 
\begin{itemize}
    \item Gradient maps (Section \ref{sec:gradient_maps}).
    \item Backpropagation (Section \ref{sec:backpropagation}).
\end{itemize}

\end{itemize}

\subsection{Handling variable-length sequences in ARMs}
\label{sec:variable_length_handling}

For completeness, we review here the crucial role of \eos to induce a valid
probability distribution over variable-length sequences of finite size.  For more details, we
refer the reader to \citet[Section 2.5]{cotterell2023formal}.  See also
\citet[Chapter 6]{eisenstein2019introduction} for a simple introduction to
language models.

\paragraph{Probability of finished vs.\ unfinished responses.}

Let us define the set of possible sequences up to size $T$ in a vocabulary $\cV$
as $\cV^{\le T} \coloneqq \cup_{t=1}^T \cV^t$. 
We use $X \in \cX$ to denote the random variable of a prompt and $Y \in \cV^{\le
T}$ to denote the random variable of a \textbf{finished} response of size at
most $T$.  We then define the probability that a sequence $\y=(y_1, \ldots,
y_\tau)$ is a finished response given $\x$ as
\[
\PP(Y=\y |X=\x) \coloneqq \PP(Y_1=y_1, \ldots, Y_\tau=y_\tau, \len(Y)=\tau |X=\x).
\]
It is the probability that the first $\tau$ tokens in $Y$ are $y_1, \ldots,
y_\tau$ \textbf{and} that the sentence is of length $0 \le \tau \le T$.

This probability should not be confused with the \textbf{marginal probability}
of an \textbf{unfinished response},
\[
\PP(Y_{\le \tau}=\y |X=\x) \coloneqq \PP(Y_1=y_1, \ldots, Y_\tau=y_\tau |X=\x).
\]
It is the probability that the first $\tau$ tokens in $Y$ are $y_1, \ldots,
y_\tau$, while $Y$ is of potentially larger length.  Formally, the event 
$\{Y_1=y_1, \ldots, Y_\tau=y_\tau, \len(Y)=\tau'\}$ 
for 
$\tau' > \tau$
is not a singleton, while the event 
$\{Y_1=y_1, \ldots, Y_\tau=y_\tau, \len(Y)=\tau\}$ 
does reduce to the singleton $\{Y=\y\}$. 
Said differently, 
$\{Y_{\le \tau} = \y\} \neq \{Y=\y\}$.

\paragraph{Explicit length handling.}

A natural way to describe finished responses of variable length would be to
define a random variable $L \in \NN$ encoding the length and to define the
distribution over fixed size sequences $Y_{1:l}|L=l$ for $l \in \NN$.
Unfortunately, in an autoregressive setting, this breaks the information flow:
we do not foresee the final length before starting to output tokens. Instead,
autoregressive models can condition the length on the past tokens, i.e., they
can decompose the probability $\PP(Y=\y |X=\x)$ as
\begin{align*}
\PP(Y=\y|X=\x) & = \PP(Y_1 = y_1, \ldots, Y_\tau=y_\tau | X=\x)
\PP(\len(Y)=\tau | X=\x, Y_1 = y_1, \ldots, Y_\tau=y_\tau) \\
& = \PP(Y_1 = y_1 |S_1 =\s_1) \ldots \PP(Y_\tau=y_\tau | S_\tau = \s_\tau)
\PP(\len(Y)=\tau | S_\tau = \s_\tau),
\end{align*}
where we used the contexts (states)
$\s_t \coloneqq \x \oplus \y_{<t}$ and $\s_1 = \x$,
and similarly for $S_t$ and $S_1$.

\paragraph{Implicit length handling using EOS.}

This decomposition suggests a simple way to encode the length in an
autoregressive model: adding a new possible output token, called \eos, such that
for any $t\geq 0$, 
\begin{equation}
\PP(Y_t = \eos | S_t = \s_t) = \PP(\len(Y) = t |S_t =\s_t).
\end{equation}
Using this approach, finished responses should now always include \eos as the
last token, and the length of a sequence now includes \eos.  More formally,
instead of defining $Y \in \cV^{\le T}$, we now define $Y \in \cY$, where $\cY =
\cup_{t=1}^T (\cV^t \times \{\eos\})$.  It should be noted that $\eos$ is just a
convenient way to encode the length and does not change the cardinality of the
set of possible sequences, since
\begin{equation}
|\cV^t \times \{\eos\}| =|\cV^t| = V^t.
\end{equation}
The advantage of the \eos approach is that we can support seamlessly both the
probability of finished responses (by appending \eos as last token) and the
marginal probability of unfinished responses (by \textbf{not} appending \eos as
last token). Indeed, assuming $y_\tau \neq \eos$, we have
\begin{equation}
\PP(Y_{\le \tau}=\y|X=\x) 
= \PP(Y_1 = y_1 |S_1 =\s_1) \ldots \PP(Y_\tau=y_\tau | S_\tau = \s_\tau).
\end{equation}

The function $q$ used to define $\pi_q$ takes a (context, next-token) pair and
returns a scalar-valued score.
Strictly speaking, the context tokens are in the vocabulary $\cV$ while the next
token is in the augmented vocabulary $\cA \coloneqq \cV \cup \{\eos\}$.
Therefore, $q \colon \cS \times \cA \to \RR$, where $\cS \coloneqq \cup_{t=1}^T
\cX \times \cV^t$.

\paragraph{Validity of the probability distribution over sequences.}

We formally give below a sufficient condition which guarantees that an ARM
defines a valid probability distribution over finished responses.
\begin{proposition}[Validity of the distribution defined by an ARM]\label{prop:valid_arm_proba}

Let $\cY \coloneqq \bigcup_{\tau=1}^T \voc^{\tau-1} \times \{\eos\}$ be the set
of valid finished responses of length at most $T$.  If $\pi_q(\eos|\s_T) = 1$
for all $\s_T \in \cX \times \voc^{T-1}$, then for all $\x \in \cX$
\begin{equation}
\sum_{\y \in \cY} p^\arm_q(\y|\x) = 1.
\end{equation}
That is, the ARM defines a valid probability distribution over finished
responses of length at most $T$.
\end{proposition}
A proof is given in Appendix \ref{proof:valid_arm_proba}.

To enforce the condition $\pi_q(\eos|\s_T) = 1$, we can define $q$ as in
\eqref{eq:q_last_token}.

\subsection{Handling variable-length sequences in EBMs}
\label{sec:variable_length_handling_ebms}

\paragraph{Probability of finished vs.\ unfinished responses.}

Defining a probability over finished responses is more straightforward with
EBMs, as we can directly parameterize $\PP(Y=\y |X=\x)$ for all possible $\y \in
\cY$.  The negative energy $R$ of the EBM is, up to a constant factor, the
log-probability of singletons events $\{Y=\y\}$ given $X=\x$, 
\begin{equation}
R(\x, \y) = \log \PP(Y=\y |X=\x) + \mathrm{const}(\x).    
\end{equation}
Marginal probabilities $\PP(Y_{\leq \tau} = \y |X=\x)$ can be computed from
$\PP(Y=\y |X=\x)$ by appropriate marginalization and similarly for the
probability of response length.

\paragraph{Validity of the probability distribution over sequences.}

An EBM defines a valid probability over finished responses of length up to $T$
by virtue of its normalization,
\[
p^{\text{EBM}}_R(\y|\x) = \frac{\exp R(\x, \y)}{\sum_{\y' \in \cY} \exp R(\x, \y')}.
\]
The denominator is well defined since $\cY$ is finite.

\subsection{Chain rule of entropy}
\label{sec:chain_rule_of_entropy}

In this section, we show how the entropy of $p$ (the sequence distribution) can
be computed in terms of the entropy of $\pi$ (the next-token distribution),
using the law of total expectation (tower rule).  This allows one to construct
an unbiased estimator of the entropy of $p$ by sampling trajectories.

\paragraph{Stochastic state transition dynamics.} \mbox{}\\

Suppose $p(\tauv|\s_1) \coloneqq \prod_{t=1}^T \pi(a_t|s_t) \mu(\s_{t+1}|\s_t, a_t)$ where $\tauv = (a_1, s_2, \dots, \s_T, a_T, s_{T+1})$. For all $\s_1$, we then have
\begin{align}
H(p(\cdot|\s_1))
&= -\EE_{\tau \sim p(\cdot|\s_1)} \left[\log p(\tau|\s_1)\right] \\
&= -\EE_{\tau \sim p(\cdot|\s_1)} \left[\sum_{t=1}^T \log \pi(A_t|S_t) + \log \mu(S_{t+1}|S_t,A_t)\right] \\
&= -\sum_{t=1}^T \EE_{S_t \sim p_t(\cdot|\s_1)} \left[E_{A_t \sim \pi(\cdot|S_t)} \left[\log \pi(A_t|S_t) + \EE_{S_{t+1} \sim \mu(\cdot|S_t,A_t)} \left[ \log \mu(S_{t+1}|S_t,A_t) \right] \right]\right] \\
&= \sum_{t=1}^T \EE_{S_t \sim p_t(\cdot|\s_1)} \left[H(\pi(\cdot|S_t))\right] + \EE_{S_t \sim p_t(\cdot|\s_1), A_t \sim \pi(\cdot|S_t)} \left[H(\mu(\cdot|S_t, A_t))\right]
\end{align}
where $p_t(S_t|\s_1)$ is the marginal probability of being in state $S_t$ at time $t$ over all possible trajectories starting from $\s_1$.

\paragraph{Deterministic state transition dynamics.} \mbox{}\\

Suppose $p(\y|\x) \coloneqq \prod_{t=1}^T \pi(y_t|\x,\y_{<t})$. For all $\x$, we now have
\begin{align}
H(p(\cdot|\x))
&= -\EE_{Y \sim p(\cdot|\x)} \left[\log p(Y|\x)\right] \\
&= -\EE_{Y \sim p(\cdot|\x)} \left[\sum_{t=1}^T \log \pi(Y_t|\x,Y_{<t})\right] \\
&= -\sum_{t=1}^T \EE_{Y_{<t} \arsim \pi(\cdot|\x)} \left[\EE_{Y_t \sim \pi(\cdot|\x,Y_{<t})} \left[\log \pi(Y_t|\x,Y_{<t})\right]\right] \\
&= \sum_{t=1}^T \EE_{Y_{<t} \arsim \pi(\cdot|\x)} \left[H(\pi(\cdot|\x, Y_{<t}))\right] \\
&= \sum_{t=1}^T H(Y_t|\x, Y_{<t})
\end{align}
where we use $\arsim$ to indicate that $Y_{<t}$ is obtained autoregressively.

Taking the expectation over $\x$, we obtain the conditional entropy
\begin{equation}
\EE_X \left[H(p(\cdot|X))\right] 
=
H(Y|X)
= 
\sum_{t=1}^T H(Y_t|X, Y_{<t}).
\end{equation}

\subsection{Chain rule of Kullback-Leibler divergence}
\label{sec:chain_rule_of_kl}

In this section, we show a similar result for the Kullback-Leibler divergence as
for the entropy.  This allows one to construct an unbiased estimator of the KL
divergence by sampling trajectories.

\paragraph{Stochastic state transition dynamics.} \mbox{}\\

Suppose 
$p(\tauv|\s_1) \coloneqq \prod_{t=1}^T \pi(a_t|s_t) \mu(\s_{t+1}|\s_t, a_t)$ 
and
$p_0(\tauv|\s_1) \coloneqq \prod_{t=1}^T \pi_0(a_t|s_t) \mu(\s_{t+1}|\s_t, a_t)$
where $\tauv = (a_1, \s_2, \dots, \s_T, a_T, \s_{T+1})$.
For all $\s_1$, we have
\begin{align} 
\mathrm{KL}(p(\cdot|\s_1), p_0(\cdot|\s_1)) 
&= \EE_{\tau \sim p(\cdot|\s_1)} \left[ \log \frac{p(\tau|\s_1)}{p_0(\tau|\s_1)} \right] \\
&= \EE_{\tau \sim p(\cdot|\s_1)} \left[ \sum_{t=1}^T \log \frac{\pi(A_t|S_t) \mu(S_{t+1}|S_t,A_t)}{\pi_0(A_t|S_t)\mu(S_{t+1}|S_t,A_t)} \right]\\ 
&= \sum_{t=1}^T \EE_{S_t \sim p_t(\cdot|\s_1)} \left[ \EE_{A_t \sim \pi(\cdot|S_t)} \left[ \log \frac{\pi(A_t|S_t)}{\pi_0(A_t|S_t)} \right] \right] \\
&= \sum_{t=1}^T \EE_{S_t \sim p_t(\cdot|\s_1)} \left[ \mathrm{KL}(\pi(\cdot|S_t), \pi_0(\cdot|S_t)) \right],
\end{align}
where $p_t(S_t|\s_1)$ is again the marginal probability of being in state $S_t$
at time $t$ over all possible trajectories starting from $\s_1$.  We see that
the state transition kernel $\mu$ cancels out.

\paragraph{Deterministic state transition dynamics.} \mbox{}\\

Similarly, suppose 
$p(\y|\x) \coloneqq \prod_{t=1}^T \pi(y_t|\x,\y_{<t})$
and
$p_0(\y|\x) \coloneqq \prod_{t=1}^T \pi_0(y_t|\x,\y_{<t})$.
For all $\x$, we have
\begin{align} 
\mathrm{KL}(p(\cdot|\x), p_0(\cdot|\x)) 
&= \EE_{Y \sim p(\cdot|\x)} \left[ \log \frac{p(Y|\x)}{p_0(Y|\x)} \right] \\
&= \EE_{Y \sim p(\cdot|\x)} \left[ \sum_{t=1}^T \log \frac{\pi(Y_t|\x, Y_{<t})}{\pi_0(Y_t|\x, Y_{<t})} \right]\\ 
&= \sum_{t=1}^T \EE_{Y_{<t} \arsim \pi(\cdot|\x)} \left[ \EE_{Y_t \sim \pi(\cdot|\x, Y_{<t})} \left[ \log \frac{\pi(Y_t|\x, Y_{<t})}{\pi_0(Y_t|\x, Y_{<t})} \right] \right] \\
&= \sum_{t=1}^T \EE_{Y_{<t} \arsim \pi(\cdot|\x)} \left[ \mathrm{KL}(\pi(\cdot|\x, Y_{<t}), \pi_0(\cdot|\x, Y_{<t})) \right] \\
&\coloneqq \sum_{t=1}^T \mathrm{KL}(\pi(Y_t|\x, Y_{<t}), \pi_0(Y_t|\x, Y_{<t}))
\end{align}
where we again use $\arsim$ to indicate that $Y_{<t}$ is obtained autoregressively.

Taking the expectation over $\x$, we obtain the conditional Kullback-Leibler divergence
\begin{equation}
\EE_X \left[\mathrm{KL}(p(\cdot|X), p_0(\cdot|X))\right] 
=
\mathrm{KL}(p(Y|X), p_0(Y|X))
= 
\sum_{t=1}^T \mathrm{KL}(\pi(Y_t|X, Y_{<t}), \pi_0(Y_t|X, Y_{<t})).
\end{equation}
Estimators based on Rao–Blackwellization are studied in \citep{amini2025better}.

\subsection{Bijection with a reference measure}
\label{sec:bijection_with_ref_measure}

Suppose we have a reference distribution
\begin{equation}
\pref(\y|\x) \coloneqq \prod_{t=1}^{|\y|} \piref(y_t|\x \oplus \y_{<t}).
\end{equation}
We then have
\begin{equation}
R_\mathrm{ref}(\x, \y)
= \log \pref(\y|\x)
= \sum_{t=1}^{|\y|} \log \piref(y_t|\x \oplus \y_{<t})
= \sum_{t=1}^{|\y|} q_\mathrm{ref}(\x \oplus \y_{<t}, y_t),
\end{equation}
where $q_\mathrm{ref}(\s, y) \coloneqq \log \piref(y|\s)$.
Suppose further that
\begin{align}
R(\x, \y) \coloneqq \sum_{t=1}^{|\y|} r(\x \oplus \y_{<t}, y_t).
\end{align}
Using the bijection for the total energy $R' \coloneqq R + R_\mathrm{ref}$, we
obtain the total logits $q'$,
\begin{equation}
q'(\s_t, y_t) \coloneqq
\begin{cases}
r(\s_t, y_t) + q_\mathrm{ref}(\s_t, y_t) &\mbox{if} ~ y_t = \eos \\
r(\s_t, y_t) + q_\mathrm{ref}(\s_t, y_t) + V_{q'}(\s_t \oplus y_t) &\mbox{if} ~ y_t \neq \eos
\end{cases}.
\end{equation}
We now define the residual logits $q$ such that $q' \coloneqq q + q_{\mathrm{ref}}$, which implies
\begin{equation}
\pi_{q'}(y_t|\s_t) \propto \piref(y_t|\s_t) \exp(q(\s_t, y_t)).
\end{equation}
Substituting $q' = q + q_\mathrm{ref}$ into the recursion above, the term
$q_\mathrm{ref}$ cancels out on both sides. The mapping between $r$ and $q$ is
therefore
\begin{equation}
q(\s_t, y_t) \coloneqq
\begin{cases}
r(\s_t, y_t) &\mbox{if} ~ y_t = \eos \\
r(\s_t, y_t) + V_{q + q_\mathrm{ref}}(\s_t \oplus y_t) &\mbox{if} ~ y_t \neq \eos
\end{cases},
\end{equation}
where
\begin{equation}
V_{q + q_\mathrm{ref}}(\s_t) = \log \EE_{Y_t \sim \piref(\cdot|\s_t)} [\exp(q(\s_t, Y_t))],
\end{equation}
which is the log-sum-exp modulated by the reference measure.

\subsection{Chain rule of probability perspective on the bijection}
\label{sec:bijection_chain_rule_perspective}

In this section, we revisit Proposition \ref{prop:bijection} from a chain rule
of probability perspective, by using the algorithm outlined in Section
\ref{sec:bijection_distributions}.  We use a fixed length $T$ for simplicity of
the exposition.  We have
\begin{align}
p(\y | \x) 
&= \frac{\exp(r(\s_1, y_1) + \dots + r(\s_T, y_T))}{Z(\x)} \\
&= \frac{\exp(r(\s_1, y_1) + \dots + r(\s_{T-1}, y_{T-1}))\exp(r(\s_T, y_T))}{Z(\x)}
\end{align}
where $Z(\x)$ is the normalization constant.

Marginalizing the last variable, we obtain
\begin{align}
p(\y_{\le T-1}|\x) 
&= \sum_{y_T \in \cV} p(\y_{<T}, y_T|\x) \\
&= \frac{1}{Z(\x)} \exp(r(\s_1, y_1) + \dots + r(\s_{T-1}, y_{T-1})) \sum_{y_T \in \cV} \exp(r(\s_T, y_T)) \\
&= \frac{1}{Z(\x)} \exp(r(\s_1, y_1) + \dots + r(\s_{T-1}, y_{T-1}))\exp(V_r(\s_T)).
\end{align}
By conditioning, we obtain
\begin{align}
\pi(y_T|\x, \y_{<T}) 
&= \frac{p(\y_{\le T}|\x)}{p(\y_{\le T-1}|\x)} \\
&= \exp(r(\s_T, y_T) - V_r(\s_T)) \\
&= \exp(q(\s_T, y_T) - V_q(\s_T))
\end{align}
where we have used that $p(\y | \x) = p(\y_{\le T} | \x)$ and
where we defined $q(\s_T, y_T) \coloneqq r(\s_T, y_T)$.

Marginalizing once again, we obtain
\begin{align}
p(\y_{\le T -2} | \x) 
&= \sum_{y_{T-1} \in \cV} p(\y_{<T-1}, y_{T-1}|\x) \\
&= \frac{1}{Z(\x)} \exp(r(\s_1, y_1) + \dots + r(\s_{T-2}, y_{T-2}))
\sum_{y_{T-1} \in \cV} \exp(r(\s_{T-1}, y_{T-1}) + V_q(\s_{T-1} \oplus y_{T-1})) \\
&= \frac{1}{Z(\x)} \exp(r(\s_1, y_1) + \dots + r(\s_{T-2}, y_{T-2}))
\exp(V_q(\s_{T-1})),
\end{align}
where we defined $q(\s_{T-1}, y_{T-1}) \coloneqq r(\s_{T-1}, y_{T-1}) +
V_q(\underbrace{\s_{T-1} \oplus y_{T-1}}_{\s_T})$.

By conditioning once again, we obtain
\begin{align}
\pi(y_{T-1}|\x, \y_{<T-1}) 
&= \frac{p(\y_{\le T-1}|\x)}{p(\y_{\le T-2}|\x)} \\
&= \exp(q(\s_{T-1}, y_{T-1}) - V_q(\s_{T-1})).
\end{align}
Repeating the process until $t=1$, we recover the mapping $\cM$
\begin{align}
q(\s_T, y_T) &\coloneqq r(\s_T, y_T) \\
q(\s_{T-1}, y_{T-1}) &\coloneqq r(\s_{T-1}, y_{T-1}) + V_q(\s_T) \\
              &~~\vdots \\
q(\s_1, y_1) &\coloneqq  r(\s_1, y_1) + V_q(\s_2).
\end{align}

\subsection{Variational perspective on the bijection}
\label{sec:bijection_variational_perspective}

In this section, we revisit Proposition \ref{prop:bijection} from a variational
perspective.  We use a fixed length $T$ for simplicity of the exposition. We
notice that
\begin{small}
\begin{align}
&A^\ebm_{\Rsub{r}}(\x) \\
=& \max_{p \in \cP(\cY|\x)} \EE_{Y \sim p(\cdot | \x)}[\Rsub{r}(\x, Y) - \log p(Y|\x)] \\
=& \max_{\pi \in \cP(\cV|\cS(\x))}
\EE_{Y_1 \sim \pi(\cdot|\x)} \left[ r(\x, Y_1) - \log \pi(Y_1|\x)
    + \dots + \EE_{Y_T \sim \pi(\cdot|\x \oplus Y_{<T})}[r(\x \oplus Y_{<T}, Y_T) - \log \pi(Y_T|\x \oplus Y_{<T})]
\right] \\
=& \max_{\substack{\pi_1 \in \cP(\cV|\cS_1(\x)))\\ \vdots\\ \pi_T \in \cP(\cV|\cS_{T-1}(\x))}}
\EE_{Y_1 \sim \pi_1(\cdot|\x)} \left[ r(\x, Y_1) - \log \pi_1(Y_1|\x)
    + \dots + \EE_{Y_T \sim \pi_T(\cdot|\x \oplus Y_{<T})}[r(\x \oplus Y_{<T}, Y_T) - \log \pi_T(Y_T|\x \oplus Y_{<T})]
\right] \\
=& \max_{\pi_1 \in \cP(\cV|\cS_1(\x))}
\EE_{Y_1 \sim \pi_1(\cdot|\x)} \left[ r(\x, Y_1) - \log \pi_1(Y_1|\x)
    + \dots + 
\max_{\pi_T \in \cP(\cV|\cS_{T-1}(\x))} 
\EE_{Y_T \sim \pi_T(\cdot|\x \oplus Y_{<T})}[r(\x \oplus Y_{<T}, Y_T) - \log \pi_T(Y_T|\x \oplus Y_{<T})]
\right].
\end{align}
\end{small}
Solving the nested maxima from right to left, it can indeed be checked 
that $A^\ebm_{\Rsub{r}}(\x) = V_q(\x)$ for all $\x \in \cX$ with $q = \cM(r)$.

\subsection{Adversarial and variational inference perspectives}
\label{sec:bijection_adversarial_perspective}

We have for all $R \in \cF(\cX \times \cY)$, $\x \in \cX$ and $\y \in \cY$,
\begin{align}
A^\ebm_{R}(\x)
&= \max_{p \in \cP(\cY|\x)} \EE_{Y \sim p(\cdot|\x)}[R(\x, Y) - \log p(Y|\x)] \\
&\ge \EE_{Y \sim p(\cdot|\x)}[R(\x, Y) - \log p(Y|\x)] \qquad \forall p \in \cP(\cY|\x).
\end{align}
In the second line, equality holds when $p = p^\ebm_R$. 

From a convex duality perspective, $R$ is the primal variable and $p$ is the
dual variable.

In particular, we have for all $R \in \cF(\cX \times \cY)$, $q \in \cF(\cS
\times \cA)$ and $\x \in \cX$,
\begin{equation}
A^\ebm_R(\x)
\ge \EE_{Y \sim p^\arm_q(\cdot|\x)}[R(\x, Y) - \log p^\arm_q(Y|\x)] \qquad \forall q \in \cF(\cS \times \cA).
\end{equation}
From Proposition \ref{prop:bijection}, equality holds when $q = \cM(r)$ and $R =
\Rsub{r}$.

We can use this perspective to derive a min-max formulation
\begin{equation}
\min_{R \in \cF(\cX \times \cY)} \EE_{(X, Y)} \left[\ell^\ebm_R(X, Y)\right]
= 
\min_{R \in \cF(\cX \times \cY)}
\max_{q \in \cF(\cS \times \cA)}
\EE_X \left[\EE_{Y' \sim p^\arm_q(\cdot|X)}[R(X, Y') - \log p^\arm_q(Y'|X)]\right] - \EE_{(X,Y)} \left[R(X, Y)\right].
\end{equation}

From a GAN or actor-critic perspective, $p_q^\arm$ is a generator (actor) and
$R$ is a discriminator (critic).

From a variational inference perspective, $A^\ebm_R(\x)$ is the log-evidence and
$(q, R) \mapsto \EE_{Y \sim p^\arm_q}[R(\x, Y) - \log p^\arm_q(Y|\x)]$ is known
as the evidence lower-bound (ELBO).  While it is a direct consequence of convex
duality, the ELBO can also be derived using Jensen's inequality or using the
non-negativity of the KL.  The distribution $p^\arm_q$ is the proposal
distribution (easy to sample from) and the distribution $p^\ebm_R$ is the target
distribution (difficult to sample from).

\subsection{Probabilistic inference perspective}
\label{sec:proba_inference_perspective}

In this section, we revisit the bijection from a probabilistic inference
perspective. Our exposition differs from \citep{levine2018reinforcement} and may
therefore bring a complementary view.
Throughout the section, we define the shorthand notation 
$\s_1 \coloneqq \x$ and $\s_t \coloneqq \x \oplus \y_{<t}$ for $t > 1$.

The conditional probability of $\y$ given $\x$ according to an EBM is
\begin{equation}
p(\y|\x) = \frac{1}{Z(\x)} \prod_{t=1}^T \psi(\s_t, y_t),
\end{equation}
where we defined the potential $\psi(\s_t, y_t) \coloneqq \exp(r(\s_t, y_t))$.

The partition function is defined as
\begin{equation}
Z(\x) \coloneqq \sum_{\y \in \cY} \prod_{t=1}^T \psi(\s_t, y_t).
\end{equation}

Using the distributivity of multiplication over addition, we can expand $Z(\x)$
as
\begin{equation}
Z(\x) = 
\sum_{y_1 \in \augvoc} \psi(\s_1, y_1) 
\sum_{y_2 \in \augvoc} \psi(\s_2, y_2)
\dots
\sum_{y_T \in \augvoc} \psi(\s_T, y_T).
\end{equation}

\paragraph{Forward and backward variables.}

Let us define the forward variables as
\begin{align}
\alpha_1(\s_1) &\coloneqq 1 \\
\alpha_t(\s_t) &\coloneqq \psi(\s_1, y_1) \dots \psi(\s_{t-1}, y_{t-1}) \qquad \forall t > 1 \\
               &=\alpha_{t-1}(\s_{t-1}) \psi(\s_{t-1}, y_{t-1}) \label{eq:alpha_recursive}.
\end{align}
They capture the accumulated value of a single specific path from the root to
the current state $\s_t$ (the past).

Likewise, let us define the backward variables as
\begin{align}
\beta_t(\s_t) &\coloneqq \sum_{y_t \in \augvoc} \psi(\s_t, y_t) \dots
\sum_{y_T \in \augvoc} \psi(\s_T, y_T) \qquad \forall t \le T \\
&= \sum_{y_t \in \augvoc} \psi(\s_t, y_t) \beta_{t+1}(\s_t \oplus y_t)
\label{eq:beta_recursive} \\
\beta_{T+1}(\cdot) &\coloneqq 1.
\end{align}
They capture the future value of being in state $\s_t$.

We then have for all $t \in [T]$
\begin{equation}
Z(\x) = \sum_{\s_t \in \cS_t(\x)} \alpha_t(\s_t) \beta_t(\s_t),
\end{equation}
where $\cS_t(\x)$ is the set of all responses $\y$ of length $t-1$ starting from
$\s_1 = \x$ (i.e., nodes at depth $t-1$ in the tree).

In particular:
\begin{itemize}
\item At the root ($t=1$), the sum collapses to a single term, $Z(\x) =
    \alpha_1(\x)\beta_1(\x) = \beta_1(\x)$.
\item At the last step ($t=T$), the sum becomes 
    $Z(\x) = \sum_{\s_T \in \cS_T(\x)} \alpha_T(\s_T) \beta_T(\s_T)$.
\end{itemize}

\paragraph{Relationship with the bijection.}

We can recover the relationship between EBMs and ARMs when examining the
recursive definition of $\beta_t$. 
Dividing \eqref{eq:beta_recursive} by $\beta_t(\s_t)$ on both sides, we obtain
\begin{equation}
\sum_{y_t \in \augvoc} \frac{\psi(\s_t, y_t) \beta_{t+1}(\s_t \oplus
y_t)}{\beta_t(\s_t)} = 1.
\end{equation}
Since the terms in the sum are positive and sum to 1, they define a
valid probability distribution over the next token $y_t$. This
distribution is exactly $\pi_q$. To see why, recall that
\begin{equation}
\pi_q(y_t | \s_t) = \exp(q(\s_t, y_t) - V_q(\s_t)).
\end{equation}
By identifying the terms, we find that
\begin{align}
V_q(\s_t) &= \log \beta_t(\s_t) \\
q(\s_t, y_t) &= \log \psi(\s_t, y_t) + \log \beta_{t+1}(\s_t \oplus y_t) \\
&= r(\s_t, y_t) + V_q(\s_t \oplus y_t).
\end{align}
We thus recover exactly the mapping $q = \cM(r)$ presented in \eqref{eq:q_from_r}.

Furthermore, we can express the joint probability of a sequence $\y$ using the
forward and backward variables. Substituting the expression for the policy
derived above into the chain rule, we obtain
\begin{align}
p^\arm_q(\y | \x) 
&= \prod_{t=1}^T \frac{\psi(\s_t, y_t) \beta_{t+1}(\s_{t+1})}{\beta_t(\s_t)} \\
&= \left( \prod_{t=1}^T \psi(\s_t, y_t) \right) \underbrace{\prod_{t=1}^T
\frac{\beta_{t+1}(\s_{t+1})}{\beta_t(\s_t)}}_{\text{telescoping product}} \\
&= \left( \prod_{t=1}^T \psi(\s_t, y_t) \right)
\frac{\beta_{T+1}(\s_{T+1})}{\beta_1(\s_1)}.
\end{align}
Using the boundary conditions $\beta_{T+1}(\s_{T+1}) = 1$ and $\beta_1(\s_1) =
Z(\x)$, we recover exactly the definition of the EBM
\begin{align}
p^\arm_q(\y | \x) &= \frac{1}{Z(\x)} \prod_{t=1}^T \psi(\s_t, y_t) \\
                  &= p^\ebm_{\Rsub{r}}(\y|\x).
\end{align}

\paragraph{Irrelevance of forward variables for sampling.}

While the forward variable $\alpha_t(\s_t)$ is essential for computing the
edge marginals $\mu_q(\s, y)$, it is
notably absent from the autoregressive generation process.
The optimal policy $\pi_q(y_t | \s_t)$ represents the conditional
probability of choosing $y_t$ given the current context $\s_t$. Using the
forward-backward decomposition of the joint probability, we have
\begin{equation}
\pi_q(y_t | \s_t) = \frac{\PP(\text{path through } \s_t \oplus
y_t)}{\PP(\text{path through } \s_t)} = \frac{\alpha_{t+1}(\s_t \oplus y_t)
\beta_{t+1}(\s_t \oplus y_t)}{\alpha_t(\s_t) \beta_t(\s_t)}.
\end{equation}
Substituting the recursive definition $\alpha_{t+1}(\s_t \oplus y_t) =
\alpha_t(\s_t) \psi(\s_t, y_t)$, we observe that the forward variable cancels
out
\begin{equation}
\pi_q(y_t | \s_t) = \frac{[\alpha_t(\s_t) \psi(\s_t, y_t)] \beta_{t+1}(\s_t
\oplus y_t)}{\alpha_t(\s_t) \beta_t(\s_t)} = \frac{\psi(\s_t, y_t)
\beta_{t+1}(\s_t \oplus y_t)}{\beta_t(\s_t)}.
\end{equation}
Intuitively, since the decision at time $t$ is conditioned on the fixed history
$\s_t$, the accumulated reward of that history ($\alpha_t$) is a constant
scaling factor that applies equally to all possible next actions, and thus does
not affect the relative probabilities. The policy depends exclusively on
the immediate reward $\psi$ and the future value $\beta$.

\paragraph{Difference with HMMs and CRFs.}

It is instructive to contrast this formulation with standard hidden Markov
models (HMMs) or linear-chain conditional random fields (CRFs). In those models,
the transition graph forms a \textbf{lattice} or \textbf{trellis} where multiple
distinct paths merge into the same state due to the Markov property. 
Consequently, the standard forward variable
$\alpha_t(\s_t)$ represents a marginal quantity obtained by \textbf{summing}
over all incoming paths. 
If we use a context size of $1$ (as in bigram HMMs), the cost of the
forward-backward algorithm is $O(T V^2)$.

In contrast, we define the state of an ARM as the unique
context $\s_t \coloneqq \x \oplus \y_{<t}$.
The underlying topology is therefore a \textbf{prefix tree}
(or trie), where every node has exactly one parent 
(except the end node, which can have several parents). 
As a result, there is a unique path from $\s_1 = \x$ to $\s_t$ and the summation
over paths typically found in the forward recursion collapses to a single
product term: there is no summation in \eqref{eq:alpha_recursive}.
The cost of dynamic programming (the algorithm is backward only) is however
$O(V^T)$.

Strictly speaking, if the
ARM has a fixed context window $W$ and the sequence length $T$ exceeds $W$, the
model becomes an order-$W$ Markov chain and the graph becomes a De Bruijn
lattice. However, in most modern LLM settings, the context window is
sufficiently large ($W \gg T$) that the non-merging tree topology is the
relevant abstraction. 
If we use a context size of $W$ (as in $n$-gram HMMs), the cost of
forward-backward is $O(T V^{W+1})$.

\subsection{Gradient maps}
\label{sec:gradient_maps}

\paragraph{Energy-based models.}

The gradient of the log-partition \wrt $R$, known as the link function, is
defined for all $\x \in \cX$ as
\begin{equation}
\nabla_R A^\ebm_R(\x) = g \in \cF(\cX \times \cY)
\end{equation}
where
\begin{equation}
g(\x', \y) \coloneqq
\begin{cases}
    p^\ebm_R(\y | \x) & \text{if}\ \x' = \x \\
    0 & \text{otherwise}
\end{cases}.
\end{equation}

\paragraph{Autoregressive models.}

By analogy, the gradient \wrt $q$ of the ``log-partition'' 
\textbf{along} the sequence $\y$ given $\x$ is
\begin{equation}
\nabla_q A^\arm_q(\x, \y) = g \in \cF(\cS \times \cA),
\end{equation}
where
\begin{equation}
g(\s_t', \cdot) \coloneqq \begin{cases}
    \pi_q(\cdot | \s_t) & \text{if}\ \s_t' = \s_t \\
    0 & \text{otherwise}
\end{cases}
\end{equation}
and where $\s_t \coloneqq \x \oplus \y_{<t}$.
Critically, this gradient map depends on the specific sequence $\y$, and the
states $\s_t$ it induces.  This highlights a fundamental geometric difference:
EBMs are defined by global normalization, while ARMs are defined by local
normalization \textbf{along} a path.

\subsection{Backpropagation}
\label{sec:backpropagation}

In this section, we derive backpropagation through $V_q(\x)$ following \citet{mensch2018differentiable}.

\paragraph{Bottom-up approach.}

Defining $V_q(\s_{T+1}) \coloneqq 0$, we run the forward pass (value
computation) from $t=T$ to $t=1$,
\begin{equation}
V_q(\s_{t}) 
= \underset{y_t \in \augvoc}{\lse} ~ r(\s_t, y_t) + V_q(\s_t \oplus y_t).
\end{equation}
where $q$ is defined from $r$ as in \eqref{eq:q_from_r}.
That is, after the forward pass is complete, we obtain $q = \cM(r)$.
Note that in \citet{mensch2018differentiable}, the topological order is
from $t=1$ to $t=T$ because the numbering of nodes is opposite to our notation.

From the identity above, the local gradients are
\begin{equation}
\frac{\partial V_q(\s_t)}{\partial V_q(\s_t \oplus y_t)}
= \frac{\partial V_q(\s_t)}{\partial r(\s_t, y_t)} 
= \pi_q(y_t|\s_t).
\end{equation}
Our goal is to compute the gradient
of $V_q(\s_1) = V_q(\x)$
\wrt the reward $r$ (edge),
\begin{equation}
\nabla_r V_q(\s_1) = G_q \in \cF(\cS \times \augvoc)
\end{equation}
where
\begin{equation}
G_q(\s_t, y_t) \coloneqq \frac{\partial V_q(\s_1)}{\partial r(\s_t, y_t)}.
\end{equation}
For convenience, we also define the accumulated gradient at state (node) $\s_t$,
\begin{equation}
\delta(\s_t) \coloneqq \frac{\partial V_q(\s_1)}{\partial V_q(\s_t)}. 
\end{equation}
Since $r(\s_t, y_t)$ influences only $V_q(\s_t)$, we have
\begin{equation}
G_q(\s_t, y_t) 
= \frac{\partial V_q(\s_1)}{\partial r(\s_t, y_t)} 
= \frac{\partial V_q(\s_1)}{\partial V_q(\s_t)} 
  \frac{\partial V_q(\s_t)}{\partial r(\s_t, y_t)}
= \delta(\s_t) \pi_q(y_t|\s_t).
\end{equation}

Similarly, 
thanks to the prefix tree structure of ARMs,
there is only one path from $\s_t$ to $\s_{t+1} = \s_t \oplus y_t$.
As a result, $V_q(\s_t \oplus y_t)$ influences only $V_q(\s_t)$ and we have
\begin{equation}
\delta(\s_{t+1}) 
= \frac{\partial V_q(\s_1)}{\partial V_q(\s_{t+1})}
= \frac{\partial V_q(\s_1)}{\partial V_q(\s_t)} \frac{\partial V_q(\s_t)}{\partial V_q(\s_{t+1})}
= \delta(\s_t) \pi_q(y_t|\s_t).
\end{equation}
Therefore, we have
\begin{equation}
G_q(\s_t, y_t)  = \delta(\s_t) \pi_q(y_t|\s_t) = \delta(\s_{t+1}).
\end{equation}
Defining $\delta(\s_1) \coloneqq 1$, 
we can compute the backward pass from $t=1$ to $T$,
\begin{equation}
G_q(\s_t, y_t)
= \delta(\s_t) \pi_q(y_t|\s_t) 
= \prod_{k=1}^t \pi_q(y_k|\s_k) = p^\arm_q(\y_{\le t} | \x).
\end{equation}
This confirms that the gradient of the log-partition function $V_q(\s_1)$ with
respect to the local reward $r(\s_t, y_t)$ for $q = \cM(r)$ is exactly the
marginal probability of the prefix $\y_{\le t}$ given $\x$.

\paragraph{Connection with forward and backward variables.}

Using the forward and backward variables from
Section \ref{sec:proba_inference_perspective}, the gradient can be equivalently written as
\begin{align}
G_q(\s_t, y_t)
&= \frac{\partial V_q(\s_1)}{\partial r(\s_t, y_t)} \\
&= \delta(\s_t) \pi_q(y_t | \s_t) \\
&= \left(\frac{\alpha_t(\s_t) \beta_t(\s_t)}{\beta_1(\s_1)}\right) 
\left(\frac{\psi(\s_t, y_t) \beta_{t+1}(\s_{t+1})}{\beta_t(\s_t)}\right) \\
&= \frac{\alpha_{t+1}(\s_{t+1}) \beta_{t+1}(\s_{t+1})}{Z(\x)},
\end{align}
where we used 
$\alpha_{t+1}(\s_{t+1}) = \alpha_t(\s_t)\psi(\s_t, y_t)$ 
and 
$\beta_1(\s_1) = Z(\x)$.

\section{Proofs}

\subsection{Lemmas}

\begin{lemma}[Stability of softargmax w.r.t. Kullback-Leibler divergence]
For all $f \colon [k] \to \RR$ and $g \colon [k] \to \RR$, we have
\begin{equation}
\KL(\softargmax(f), \softargmax(g)) \le 2 \|f - g\|_\infty
\end{equation}
where $\|h\|_\infty \coloneqq \max_{j \in [k]} |h(j)|$.
\label{lemma:kl_bound}
\end{lemma}
\begin{proof}
Let us define $p_f \coloneqq \softargmax(f)$ and $p_g \coloneqq \softargmax(g)$.
We have
\begin{align}
\KL(p_f, p_g) 
&= \sum_{i \in [k]} p_f(i) \log \frac{p_f(i)}{p_g(i)} \\
&= \sum_{i \in [k]} p_f(i) ((f(i) - \log Z_f) - (g(i) - \log Z_g)) \\
&= \EE_{i \sim p_f} [f(i) - g(i)] - (\log Z_f - \log Z_g)
\end{align}
where 
$Z_f \coloneqq \sum_{j \in [k]} \exp(f(j))$
and
$Z_g \coloneqq \sum_{j \in [k]} \exp(g(j))$.

Let us define $\varepsilon \coloneqq \|f - g\|_\infty$.

Using $f(j) - g(j) \le \varepsilon$ for all $j \in [k]$, we obtain
\begin{align}
Z_f 
&= \sum_{j \in [k]} \exp(f(j)) \\
&= \sum_{j \in [k]} \exp(g(j) + (f(j) - g(j))) \\
&\le \sum_{j \in [k]} \exp(g(j) + \varepsilon) \\
&= \exp(\varepsilon) \sum_{j \in [k]} \exp(g(j)) \\
&= \exp(\varepsilon) Z_g
\end{align}
and therefore
\begin{equation}
\log Z_f \le \log Z_g + \varepsilon.
\end{equation}
Applying the same reasoning
with $g(j) - f(j) \le \varepsilon$ for all $j \in [k]$, we obtain
\begin{equation}
\log Z_g \le \log Z_f + \varepsilon.
\end{equation}
Therefore, we obtain the tight bound
\begin{equation}
|\log Z_f - \log Z_g| \le \varepsilon.
\end{equation}
We also have
\begin{equation}
\EE_{i \sim p_f} [f(i) - g(i)]
\le \EE_{i \sim p_f} [\varepsilon]
= \varepsilon.
\end{equation}
Overall, we therefore obtain
\begin{equation}
\KL(p_f, p_g) \le 2 \varepsilon = 2 \|f - g\|_\infty.
\end{equation}

\end{proof}

\subsection{Proof of bijection (Proposition \ref{prop:bijection})}
\label{proof:bijection}

\paragraph{Mapping from ARM to EBM.}

Given a context $\x \in \cX$, the probability according to an ARM of a response
$\y = (y_1, \ldots, y_\tau) \in \cY$, with $y_\tau = \eos$, decomposes as 
\begin{align}
p^\arm_q(\y|\x)
&=\pi_q(y_1|\s_1) \pi_q(y_2|\s_2) 
\dots \pi_q(y_{\tau}|\s_{\tau}).
\end{align}
In the above, we defined the intermediate contexts as
\begin{equation}
\s_t \coloneqq
\begin{cases}
\x &\mbox{if } t = 1 \\
\x \oplus \y_{<t} &\mbox{if } t > 1
\end{cases},
\end{equation}
the conditional probabilities $\pi_q$ as
\begin{align}
    \pi_{q}(y_1|\s_1) &\coloneqq \exp(q(\s_1, y_1) - V_q(\s_1)) \\
    \pi_q(y_2|\s_2) &\coloneqq \exp(q(\s_2, y_2) - V_q(\s_2)) \\
                     &~~\vdots \\
    \pi_q(\eos|\s_{\tau}) &\coloneqq \exp(q(\s_{\tau}, \eos) - V_q(\s_{\tau})) 
\end{align}
and the (local) log-partitions as
\begin{equation}
V_q(\s_t) \coloneqq \log \sum_{y_t\in \augvoc} \exp q(\s_t, y_t).
\end{equation}
We then get
\begin{align}
p^\arm_q(\y|\x)
&= \exp(q(\s_1, y_1) + q(\s_2, y_2) + \dots + q(\s_{\tau}, \eos) 
- V_q(\s_1) - V_q(\s_2) - \dots - V_q(\s_{\tau}))
\\
&= \exp(r(\s_1, y_1) + r(\s_2, y_2) + \dots + r(\s_{\tau}, \eos) 
- V_q(\s_1)) \\
&= p^\ebm_{\Rsub{r}}(\y|\x),
\end{align}
where $\Rsub{r}$ is defined in \eqref{eq:R_decomposition} and where we defined
\begin{align}
r(\s_1, y_1) &\coloneqq q(\s_1, y_1) - V_q(\s_2) \\
              &~~\vdots \\
r(\s_{\tau-1}, y_{\tau-1}) &\coloneqq q(\s_{\tau-1}, y_{\tau-1}) - V_q(\s_{\tau}) \\
r(\s_\tau, \eos) & \coloneqq q(\s_\tau, \eos).
\end{align}

The value $V_q(\s_1) = V_q(\x)$ is a valid log-partition function for
$p^\ebm_{\Rsub{r}}$, provided that $p^\arm_q$ is a valid distribution.  The latter is
ensured if $\pi_q(\eos|\s_T) = 1$. This can be achieved by fixing 
\begin{equation}
q(\s_T, y_T) \coloneqq
\begin{cases}
0 &\mbox{if } y_T = \eos \\
-\infty &\mbox{if } y_T \neq \eos
\end{cases}
\end{equation}
for any $\s_T \in \cS_T$, where $\cS_T \coloneqq \cX \times (\augvoc \setminus
\{\eos\})^{T-1}$.

The mapping above is valid for any $\tau$ and any $y_1, \ldots, y_\tau$ (with
$y_\tau = \eos$ in all cases) so it can be summarized simply as
\begin{equation}
r(\s, y) \coloneqq
\begin{cases}
q(\s, y) &\mbox{if} ~ y = \eos \\
q(\s, y) - V_q(\s \oplus y) &\mbox{if} ~ y \neq \eos
\end{cases},
\end{equation}
for $\s \in \cS$ and $y \in \augvoc$.
This is the mapping $r = \cM^{-1}(q)$.

\paragraph{Mapping from EBM to ARM.}

The reverse mapping is obtained by back-substitution
\begin{align}
q(\s_\tau, \eos) &\coloneqq r(\s_\tau, \eos) \\
q(\s_{\tau-1}, y_{\tau-1}) &\coloneqq r(\s_{\tau-1}, y_{\tau-1}) + V_q(\s_T) \\
              &~~\vdots \\
q(\s_1, y_1) &\coloneqq  r(\s_1, y_1) + V_q(\s_2).
\end{align}
This is the mapping $q = \cM(r)$.

The existence of the explicit inverse $\cM^{-1}$ ensures that $\cM$ is a bijective mapping.

\subsection{Proof of same minima (Proposition \ref{prop:same_minima})}
\label{proof:same_minima}

Because we can always define $r$ from $R$ using
\begin{equation}
r(\s_t, y_t) \coloneqq
\begin{cases}
0 &\mbox{if} ~ y_t \neq \eos  \\
R(\x, \y) &\mbox{if} ~ y_t = \eos \\
\end{cases},
\label{eq:R_from_r}
\end{equation}
we have
\begin{equation}
\min_{r \in \cF(\cS \times \augvoc)} 
\EE_{(X,Y) \sim \rho} \left[ \ell^\ebm_{\Rsub{r}}(X, Y) \right]
=
\min_{R \in \cF(\cX \times \cY)} 
\EE_{(X,Y) \sim \rho} \left[ \ell^\ebm_{R}(Y|X) \right].
\end{equation}
Because the mapping $\cM$ is bijective (and therefore surjective), we have
\begin{align}
\min_{r \in \cF(\cS \times \augvoc)} 
\EE_{(X,Y) \sim \rho} \left[ \ell^\ebm_{\Rsub{r}}(X, Y)\right]
&=
\min_{r \in \cF(\cS \times \augvoc)} 
\EE_{(X,Y) \sim \rho} \left[ -\log p^\ebm_{\Rsub{r}}(Y|X\right]) \\
&=
\min_{r \in \cF(\cS \times \augvoc)} 
\EE_{(X,Y) \sim \rho} \left[ -\log p^\arm_{\cM(r)}(Y|X)\right] \\
&=
\min_{q \in \cF(\cS \times \augvoc)} 
\EE_{(X,Y) \sim \rho} \left[ -\log p^\arm_q(Y|X)\right] \\
&=
\min_{q \in \cF(\cS \times \augvoc)} 
\EE_{(X,Y) \sim \rho} \left[ \ell^\arm_q(X, Y)\right].
\end{align}

\subsection{Proof of KL bound (Proposition \ref{prop:kl_bound})}
\label{proof:kl_bound}

From Lemma \ref{lemma:kl_bound}, for all $\s_t \in \cS$, we have
\begin{equation}
\KL(\pi_q(\cdot|\s_t), \pi_{q^\star}(\cdot|\s_t))
\le
2 \|q^\star(\s_t, \cdot) - q(\s_t, \cdot)\|_\infty.
\end{equation}
This bounds the token-level distribution error.

We then have, for all $\x \in \cX$,
\begin{align}
\KL(p^\arm_q(\cdot|\x), p^\arm_{q^\star}(\cdot|\x))
&= \sum_{t=1}^T
\EE_{S_t} \KL(\pi_{q^\star}(\cdot|S_t), \pi_q(\cdot|S_t)) \\
&\le 
2 \sum_{t=1}^T
\EE_{S_t} \|q^\star(S_t, \cdot) - q(S_t, \cdot)\|_\infty \\
&\le 2T \max_{\s \in \cS(\x)} \|q^\star(\s, \cdot) - q(\s, \cdot)\|_\infty,
\end{align}
where $S_t$ above is defined by $S_t \coloneqq \x \oplus Y_{<t}$, where $Y_{<t}
\arsim \pi_{q^\star}(\cdot|\x)$.  See also Appendix \ref{sec:chain_rule_of_kl}
for more details.

This bounds the sequence-level distribution error.

Finally, using Proposition \ref{prop:bijection}, we obtain
\begin{align}
\KL(p^\arm_q(\cdot|\x), p^\ebm_{\Rsub{r}}(\cdot|\x))
&= \KL(p^\arm_{\cM(r)}(\cdot|\x), p^\arm_q(\cdot|\x)) \\
&\le 2T \max_{\s \in \cS} \|\cM(r)(\s, \cdot) - q(\s, \cdot)\|_\infty.
\end{align}

\subsection{Proof that an ARM defines a valid probability distribution
(Proposition \ref{prop:valid_arm_proba})}
\label{proof:valid_arm_proba}

By definition of $\cY$, we have 
\begin{equation}
\sum_{\y \in \cY} p^\arm_q(\y|\x) = \sum_{t=1}^T 
\sum_{\y_{<t} \in \voc^{t-1}} p^\arm_q(\y_{<t} \oplus \eos | \x) = \sum_{t=1}^T l_t
\end{equation} 
where
\[
l_t \coloneqq \sum_{\y_{<t} \in \voc^{t-1}} p^\arm_q(\y_{<t} \oplus \eos | \x),
\]
can be interpreted as the probability that a response has length $t$.
We can decompose $l_t$ for any $1 \leq t \leq T$ as follows
\begin{align}
l_t 
& = \sum_{\y_{<t} \in \voc^{t-1}} \left[ \prod_{k=1}^{t-1} \pi_q(y_k|\s_k) \right] \pi_q(\eos|\s_t) \\
& \stackrel{(i)}{=} 
\sum_{\y_{<t} \in \voc^{t-1}} \left[ \prod_{k=1}^{t-1} \pi_q(y_k|\s_k) \right]\left(1 - \sum_{y_t \in \cV}\pi_q(y_t|\s_t)\right) \\
& = M_t - M_{t+1},
\end{align}
where $M_1\coloneqq1$ and for $t>1$,
\[
M_t \coloneqq \sum_{\y_{<t} \in \voc^{t-1}} \prod_{k=1}^{t-1} \pi_q(y_k|\s_k)
= \sum_{\y_{<t} \in \voc^{t-1}} p_q^\arm(\y_{<t}|\x) 
\]
can be interpreted as the probability mass of sequences that have \textit{not}
terminated strictly before step $t$.  In the derivations above, the equality in
$(i)$ stems from the definition of the conditional probabilities through a
$\softargmax$. Namely, since $\pi_q(\cdot |\s_t) = \softargmax (q(\s_t,
\cdot))$, we have that
\begin{align}
&\sum_{y_t \in \voc \cup \{\eos\}} \pi_q(y_t|\s_t) = \sum_{y_t \in \voc \cup \{\eos\}} \softargmax(q(\s_t, y_t)) = 1 \\
\implies &\sum_{y_t \in \voc}\pi_q(y_t|\s_t) = 1 - \pi_q(\eos|\s_t).
\end{align}
We then obtain by telescoping the $M_t$ terms, 
\begin{equation}
\sum_{t=1}^T l_t = M_1 - M_{T+1} = 1 - M_{T+1}.
\end{equation}
By assumption, $\pi_q(\eos|\s_T) = 1$, which implies $\pi_q(y_T|\s_T) = 0$ for
any $y_T \in \voc$. Consequently, no mass can continue beyond $T$, therefore $M_{T+1} =
0$ and we have shown the claim.

\section{Details on numerical illustrations} \label{app:exp_details}

\subsection{Setup}

\paragraph{Architecture.}

Both EBMs and ARMs use as a backbone a pre-layer norm Transformer with a scaled
dot-product multihead attention, a positional embedder and a ReLU activation.
We introduce the following notation:
\begin{itemize}[nosep]
    \item Size of the vocabulary: $V$,
    \item Length of the sequences: $T$,
    \item Number of blocks: $N$,
    \item Embedding dimension: $D$,
    \item Number of heads in 
    the attention layer: $H$,
    \item Hidden dimension in the MLP: $F$.
\end{itemize}
The backbone \texttt{Transformer} is summarized below in abbreviated Flax
code~\citep{flax2020github}.  It takes a sequence \texttt{x\_T}$\in \cV^T$ and
outputs a representation \texttt{x\_TxD}$\in (\RR^V)^T$.

\begin{lstlisting}[language=Python]
class TransformerBlock(H, D, F):
    def __call__(x_TxD, mask):
        y_TxD = LayerNorm()(x_TxD)
        y_TxD = MultiHeadAttention(H, D)(y_TxD, mask)
        x_TxD = y_TxD + x_TxD
    
        y_TxD = LayerNorm()(x_TxD)
        y_TxF = Linear(F)(y_TxV)
        y_TxF = relu(y_TxF)
        y_TxD = Linear(D)(y_TxF)
        x_TxD = y_TxD + x_TxD
        return x_TxD

class Transformer(V, T, N, H, D, F):
    def __call__(x_T, mask):
        x_TxD = Embedder(V, D)(x_T)
        y_TxD = Embedder(T, D)([0, ...,L])
        x_TxD = x_TxD + y_TxD
        for i in range(N):
            TransformerBlock(H, D, F)(x_TxD, mask)
        return x_TxD
\end{lstlisting}
For the ARM, the autoregressive dependency pattern is encoded through a
\texttt{mask} applied to the attention matrix. The EBM does not use any mask.
The ARM projects the embedded tokens into a space of size $V$ to define one
logit per token.  The EBM takes the average representation of the sequence along
the sequence indexes and applies a final layer on the $D$ dimensional
representation of the whole sequence to get a single score.

\begin{lstlisting}[language=Python]
class AutoRegressiveModel((V, T, N, H, D F):
    def __call__(x_T):
        mask = make_causal_mask(x)
        x_TxD = Transformer(V, T, N, H, D, F)(x_TxD, mask)
        x_TxV = Linear(V)(x_TxD)
        return x_TxV

class EnergyBasedModel(V, T, N, H, D, F):
    def __call__(x_T):
        x_TxD = Transformer(V, T, N, H, D, F)(x_TxD, None)
        x_D = mean(x_TxD, axis=0)
        x = Linear(1)(x_D)
        return x
\end{lstlisting}

All sequences start with a \bos index.  The \bos index is part of the embedder
vocabulary but not part of the head (last linear transformation) of the ARM,
since sentences can only end and not start again.
 
\paragraph{Objective function.}

We compute the exact expected risk \eqref{eq:expected_risk_arm} and
\eqref{eq:expected_risk_ebm} by summing over all possible sequences and
parameterizing the data distribution with a probability mass function $\rho$ of
our choice. For simplicity, we use a fixed length $T$. Concretely, using the EBM
as an example, we compute the logits $R(\x, \y)$ for all possible sequences
$(\x, \y)$ of length $T$ and then compute the loss using
\[
\cL^{\ebm}(R) 
= \sum_{(\x, \y) \in \cV^T} \rho(\x, \y) \ell_R^{\ebm}(\x, \y).
\]
For the data distribution $\rho$, we tried two settings.
\begin{enumerate}[nosep]
    \item A Zipfian distribution over the set of all $V^T$ sequences. This gives
        a distribution over all possible sequences with a large tail.
    \item A normal-softargmax distribution, i.e., 
    $\rho = \softargmax(z/t)$ for $z \sim \mathcal{N}(0, 1)$ a random normal vector,
    and $t>0$ a temperature parameter. For small $t$, this gives a vector of probability 
    close to a one-hot vector among all possible sequences.
\end{enumerate}

We briefly mentioned in the main text that $\cL^\star = -\EE_{(X,Y) \sim \rho}
\log \rho(\x, \y)$ was the minimum of $\cL^\ebm(R)$. To verify this, it suffices
to observe that
\[
\cL^{\ebm}(R) - \cL^\star = \KL(\rho, \softargmax(R)) \geq 0.
\] 

\paragraph{Optimization.}

We use the Adam optimizer~\citep{kingma2014adam} with a trapezoidal schedule
(warm-up of $100$ steps, constant learning rate $\eta$, decay of 100 steps) with
a learning rate $\eta$ searched in $[3\cdot10^{-3}, 1\cdot10^{-3}, 10^{-4}]$. We
observed that the best learning rate (measured in terms of minimal area under
the training curve) for EBMs matched the best learning rate for ARMs.
Therefore, in the plots below, we simply choose the best learning rate for the
EBM and use the same for the ARM. 

\subsection{Configurations.}
We consider the following sequence spaces.
\begin{enumerate}[nosep]
    \item $V=8, T=4$, leading to $4\,096$ sequences.
    \item $V=4, T=8$, leading to $65\,536$ sequences.
\end{enumerate}

For these sequence spaces, we consider the following architectures 
\begin{enumerate}[nosep]
    \item $2$-layer Transformer with $N=2, H=2, D=16, V=16$.
                For  
                $V=8, T=4$ 
                or 
                $V=4, T=8$,
                leading to to $3\,633$ parameters.
    \item $3$-layer Transformer with $N=3, H=4, D=64, F=64$.
                For  
                $V=8, T=4$ 
                or 
                $V=4, T=8$,
                leading to $76\,609$ parameters.
\end{enumerate}

The $2$-layer Transformer is under-parameterized for $V=8, T=4$ and $V=4, T=8$,
while the $3$-layer Transformer is over-parameterized in those cases.

\subsection{Metrics illustrated}

\begin{figure}
    \centering
    \includegraphics[width=0.6\linewidth]{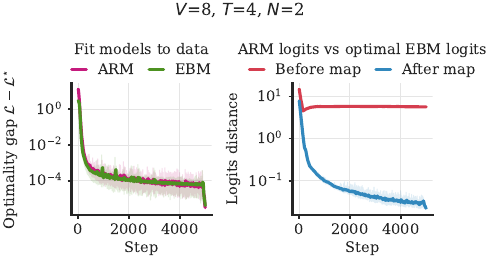}
    \includegraphics[width=0.6\linewidth]{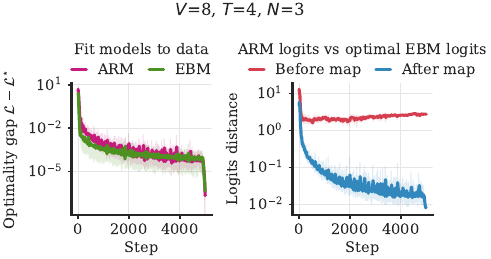}
    \caption{Loss convergence and logits distances for different Transformer sizes.}
    \label{fig:cvg_logit_dist_N_vary}
\end{figure}

\begin{figure}[p]
    \centering
    \includegraphics[width=0.6\linewidth]{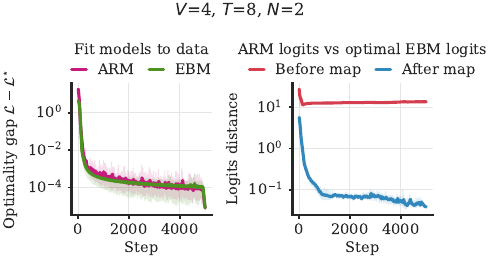}
    \includegraphics[width=0.6\linewidth]{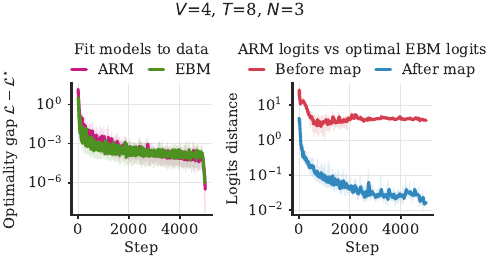}
    \caption{Loss convergence and logits distances for different Transformer sizes in the case $T>V$.}
    \label{fig:cvg_logit_dist_T_much_gt_V}
\end{figure}

To illustrate Proposition~\ref{prop:same_minima}, we plot the losses along the
training of these Transformers. Namely, in the left panel of
Figures~\ref{fig:ebm_arm_bij_illus},~\ref{fig:cvg_logit_dist_N_vary}
and~\ref{fig:cvg_logit_dist_T_much_gt_V}, we plot 
\[
\cL^\ebm(R(w_k)) - \cL^\star, \quad \text{and} \quad \cL^\arm(q(w_k)) - \cL^\star,
\]
where $w_k^\arm$ and $w_k^\ebm$ are the weights of, respectively, the ARM and
EBM, at iteration $k$ of the training. 

To illustrate our bijection result, after having optimized the EBM, we save the
corresponding logits $R^\star(\x, \y)$, and we then compare the logits $q_k(\x,
\y)$ of the ARM at iteration $k$ of its training, against the optimal EBM
logits. We perform two comparisons. 

The first one consists in simply summing the logits of the ARM along the
sequence to get a scalar $\bar q_k(\x, \y) = \sum_{t=1}^T q_k(\s_t, y_t)$ with
$\s_0 = \x = \bos$ in our case, and $\s_t, y_t$ defined as in the main text.
This comparison amounts to think of the logits of the ARM as a score in itself
like in the definition of the EBM model.  We then center the scores of both the
ARM and the EBM model (i.e., substract the mean score among all possible
sequences), and measure the maximal absolute difference between them. Formally,
we plot 
\[
\|c(\bar q_k) - c(R^\star)\|_\infty,
\]
where $c(\z) = \z - \langle\z, \ones\rangle\ones$.
In Figures~\ref{fig:ebm_arm_bij_illus},~\ref{fig:cvg_logit_dist_N_vary}
and~\ref{fig:cvg_logit_dist_T_much_gt_V}, we call this measurement of the logits
distance ``Before mapping''. 

The second comparison consists in computing $q^\star = \cM(r^\star)$ for 
\[
r^\star(\s_t, y_t) = \begin{cases}
    0 & \ \text{if} \ y_t \neq \eos \\
    R^\star(\x, \y) & \ \text{if} \ y_t = \eos
\end{cases}.
\]
This amounts to computing the ARM logits corresponding to the optimal EBM model
via the mapping $\cM$ used in Proposition~\ref{prop:bijection}.  We then center
along the last axis the logits of the ARM at iteration $k$ and the logits of
the EBM mapped in AR form. Finally, we plot their maximal absolute
elementwise distance, i.e., we plot 
\[
\|c_{-1}(q_k) - c_{-1}(\cM(r^\star))\|_\infty,
\]
where $c_{-1}$ denotes centering along the last axis.  In
Figures~\ref{fig:ebm_arm_bij_illus},~\ref{fig:cvg_logit_dist_N_vary}
and~\ref{fig:cvg_logit_dist_T_much_gt_V}, we call this measurement the logits
distance ``After mapping''. 

In both cases, the centering is necessary to take care of the fact that, up to a
constant factor, logits may define the same probability distribution through the
$\softargmax$ operator \citep[Remark 4.2]{edpbook}.  Centering lets us discover,
whether, up to some constant factor, the logits of the ARM converge to the
logits predetermined by the EBM after an appropriate mapping.

\subsection{Convergence and logits distance comparisons}

Figure~\ref{fig:ebm_arm_bij_illus} illustrates the metrics explained above in
the case $V=8$, $T=4$, for the $2$-layer Transformer presented above.

We present some additional results here. 
\begin{itemize}[nosep]
    \item In Figure~\ref{fig:cvg_logit_dist_N_vary}, we vary the size of the
        Transformer and observe the same conclusions as in the main text. 
    \item In Figure~\ref{fig:cvg_logit_dist_T_much_gt_V}, we also try the case
        $T>V$, namely, $T=8, V=4$: same conclusions apply as in the main text.
\end{itemize}
The convergence in logits is challenged when using a different data distribution
as explained in the next paragraph.

\subsection{Bounding KL divergence by logits distance}

\begin{figure}[p]
    \centering
    \includegraphics[width=\linewidth]{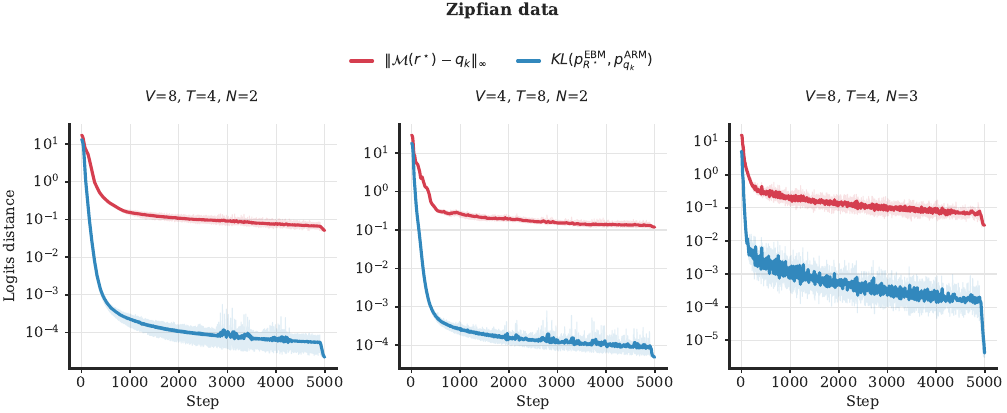}
    \includegraphics[width=\linewidth]{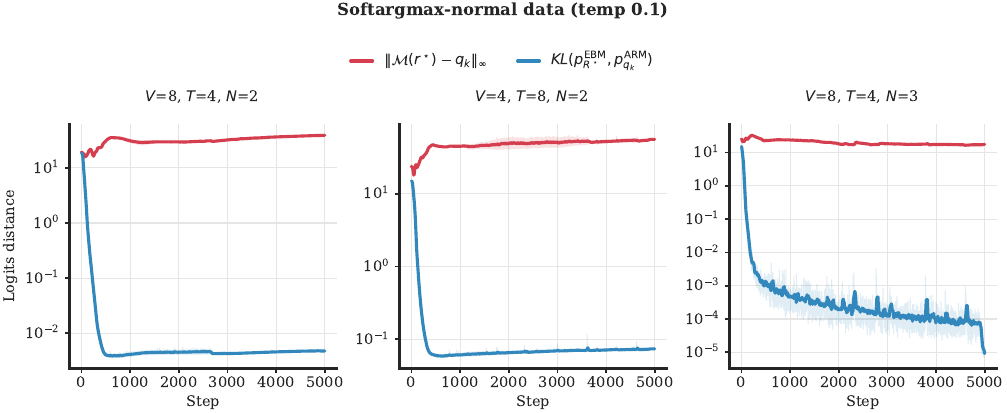}
    \caption{Comparing KL divergence and logits distance in infinity norm.}
    \label{fig:kl_vs_inf_norm}
\end{figure}

In Figure~\ref{fig:kl_vs_inf_norm}, we compare KL divergence and distance
between logits in infinity norm. Namely, as in the previous experiments, we
compute the optimal logits for the EBM and then compare the logits of the ARM
with the logits of the EBM mapped into ARM form. We use either the maximal
absolute distance between logits, or we compare the KL distance between the
probability distributions that the EBM and ARM define. We consider several setups ($V>T$,
$T<V$, $N=2$ or $3$) with different definitions of the target distribution
$\rho$.

Proposition~\ref{prop:kl_bound} stated that the KL should be upper bounded by
the maximal absolute distance between logits. This is empirically confirmed in
Figure~\ref{fig:kl_vs_inf_norm}. 

Note that when we define the target distribution as a $\softargmax$ of a normal
vector (bottom panels in Figure~\ref{fig:kl_vs_inf_norm}), only the KL converges
to 0, not the distance in infinity norm. In this case, the target distribution
is extremely peaked towards a single sequence.  So even if the distributions
match after mapping the logits to the same vertex in the simplex of all
sequences, the logits themselves may be far from each other.  Another way to see
that, is that the inverse mapping of the $\softargmax$ is not Lipschitz
continuous around a vertex of the simplex.

\clearpage

\subsection{Optimization dynamics}

\paragraph{Effect of learning rate.}

\begin{figure}[h!]
    \centering
    \includegraphics[width=0.50\textwidth]{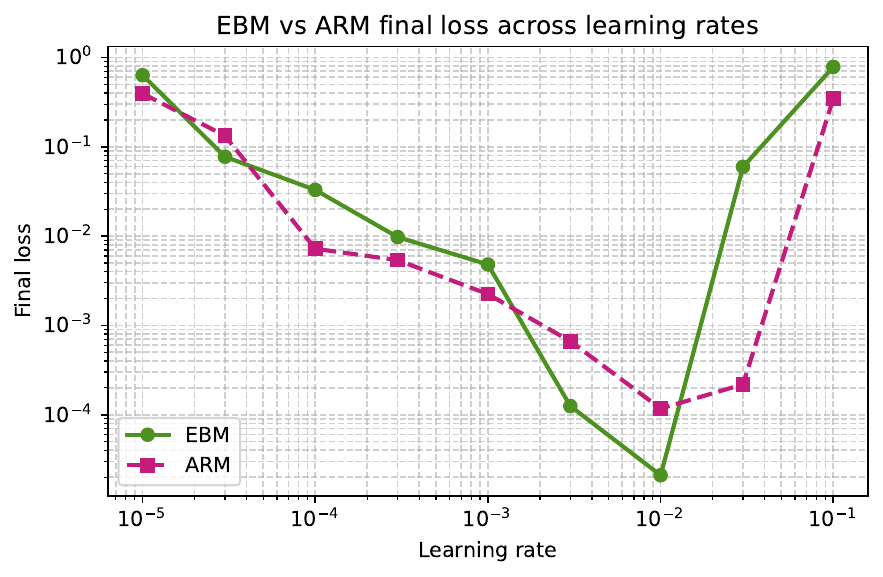}
    \caption{Final loss across learning rates. Both EBMs and ARMs achieve optimal performance at the same learning rate but ARMs are slightly more robust to learning rate specification.}
    \label{fig:learning_rate_sweep}
\end{figure}

\begin{figure}[h!]
    \centering
    \includegraphics[width=0.75\textwidth]{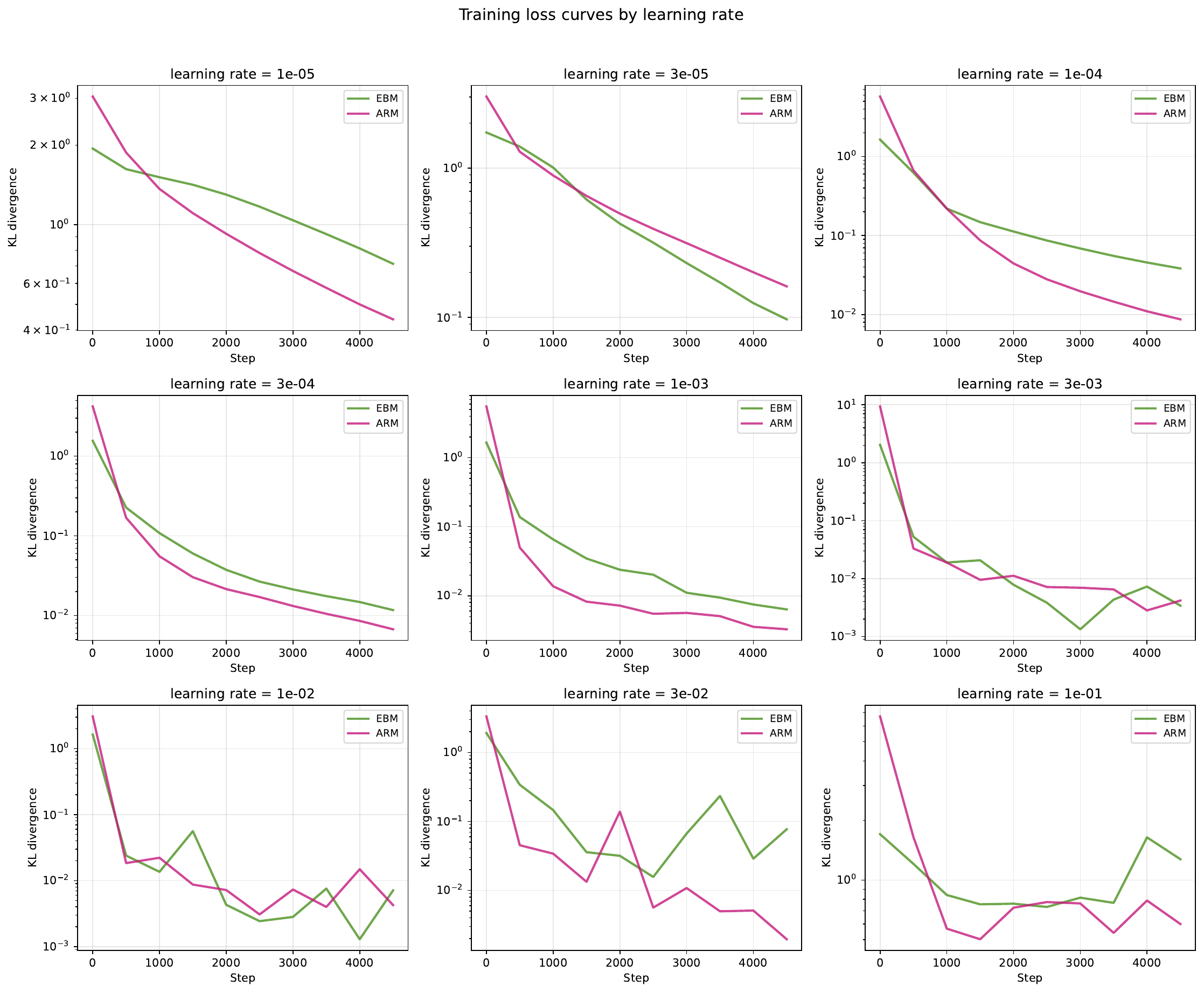}
    \caption{Training loss curves per learning rate. ARMs are more robust to learning rate specification.}
    \label{fig:training_loss_curves_by_learning_rate}
\end{figure}

In the previous experiments, we reported the results with the best learning rate. In Figure \ref{fig:learning_rate_sweep} and Figure \ref{fig:training_loss_curves_by_learning_rate}, we compare EBMs and ARMs across different learning rates. In this experiment, both EBMs and ARMs achieve optimal performance at the same learning rate but ARMs are slightly more robust to learning rate specification.

\paragraph{Gradient noise analysis.}

\begin{figure}[h]
    \centering
    \includegraphics[width=0.95\textwidth]{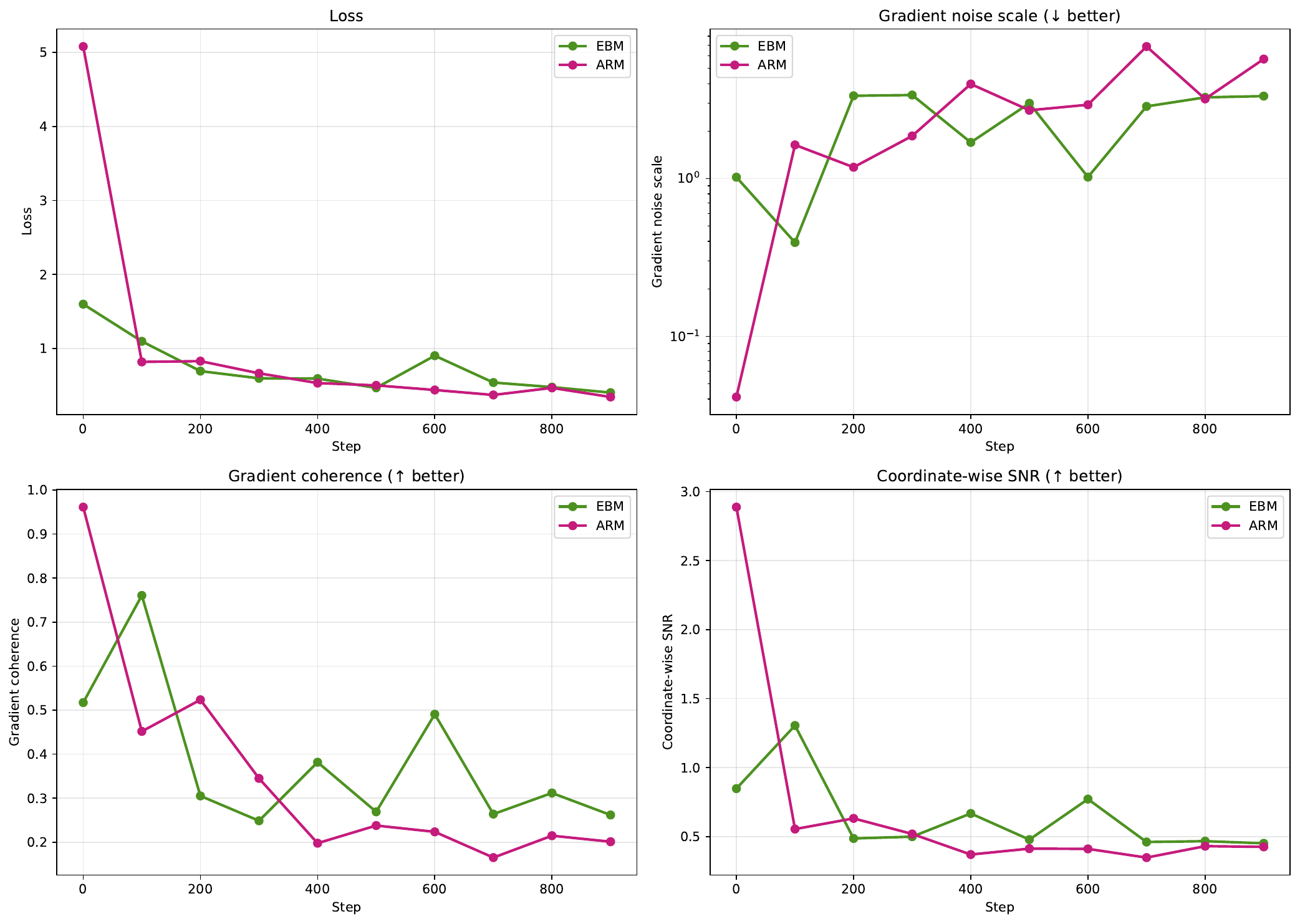}
    \caption{Gradient noise analysis. The three metrics are
    $\text{GNS} \coloneqq \frac{\mathbb{E}_B\left[\|g_B - g\|^2\right]}{\|g\|^2}$,
    $\text{Coherence} \coloneqq \frac{\|\mathbb{E}_B[g_B]\|^2}{\mathbb{E}_B\left[\|g_B\|^2\right]}$,
    and $\text{SNR} \coloneqq \frac{1}{D} \sum_{i=1}^{D} \frac{|\mu_i|}{\sigma_i}$,
    where $g_B$ is the mini-batch gradient,
    $\mu_i \coloneqq \mathbb{E}_B[(g_B)_i]$, 
    $\sigma_i \coloneqq \sqrt{\mathrm{Var}_B[(g_B)_i]}$
    and $D$ is the number of parameters.   
    EBMs and ARMs have similar sensitivity to gradient noise.
    }
    \label{fig:gradient_noise}
\end{figure}

In Figure \ref{fig:gradient_noise}, we analyzed gradient noise using three metrics: gradient noise scale (GNS), gradient coherence and coordinate-wise signal-to-noise (SNR) ratio (see caption for their definitions. In this experiment, EBMs and ARMs are similarly sensitive to noise. This confirms once again one of the main messages of our paper: ARMs have the ability to behave very similarly to EBMs (while being much easier to train and sample from).

\end{document}